\documentclass[journal]{IEEEtran}

\usepackage{cite}
\usepackage[utf8]{inputenc}
\usepackage{csquotes}
\usepackage[T1]{fontenc}
\usepackage{tabularx}
\usepackage{booktabs}
\usepackage{multirow}
\usepackage{caption}
\usepackage{subcaption}
\usepackage{algpseudocode}
\usepackage{algorithm}
\usepackage{comment}
\usepackage[scaled]{helvet}
\usepackage{booktabs}
\usepackage{hyperref}
\usepackage[utf8]{inputenc}
\usepackage{textcomp}
\usepackage{amsmath}
\usepackage{mathrsfs}
\usepackage[english]{babel}
\usepackage{collcell}
\usepackage{array, hhline}
\usepackage{pgf}
\usepackage{float}
\usepackage{subcaption}
\usepackage{amssymb}
\usepackage{xtab,afterpage}
\usepackage{xspace}
\usepackage{multicol}
\usepackage{colortbl}
\usepackage{makecell}
\usepackage{comment}
\usepackage{tikz}
\usepackage{pgfplots}

\usepackage{seqsplit}

\begin{document}

\title{RDD4D: 4D Attention-Guided Road Damage Detection And Classification}

\author{Asma~Alkalbani,
        Muhammad~Saqib,
        Ahmed~Salim~Alrawahi,
        Abbas~Anwar,
        Chandarnath~Adak,
        Saeed~Anwar

\thanks{Asma~Alkalbani, Ahmed~Alrawahi are with the Department of Information Technology, College of Computing and Information Sciences, University of Technology and Applied Sciences, Ibri, 516, Oman.}
\thanks{Muhammad~Saqib is with NCMI, CSIRO 
 and the University of Technology 
Sydney, Australia}
\thanks{Abbas~Anwar is a research assistant with FutureDataMinds}
\thanks{Chandarnath~Adak is with Indian Institute of Technology Patna, India}
\thanks{Saeed~Anwar is with the School of Computing, Australian National University, Australia}
\thanks{e-mail:~asmam.alkalbani@utas.edu.om~(Corresponding~Author),
muhammad.saqib@csiro.au,
ahmed.alrawahi@utas.edu.om, abbas.anwar@futuredataminds.com, 
chandranath@iitp.ac.in,
saeed.anwar@anu.edu.au}
}

\IEEEtitleabstractindextext{%
\begin{abstract}
Road damage detection and assessment are crucial components of infrastructure maintenance. However, current methods often struggle with detecting multiple types of road damage in a single image, particularly at varying scales. This is due to the lack of road datasets with various damage types having varying scales. To overcome this deficiency, first, we present a novel dataset called Diverse Road Damage Dataset (DRDD) for road damage detection that captures the diverse road damage types in individual images, addressing a crucial gap in existing datasets. Then, we provide our model, RDD4D, that exploits Attention4D blocks, enabling better feature refinement across multiple scales. The Attention4D module processes feature maps through an attention mechanism combining positional encoding and "Talking Head" components to capture local and global contextual information. In our comprehensive experimental analysis comparing various state-of-the-art models on our proposed, our enhanced model demonstrated superior performance in detecting large-sized road cracks with an Average Precision (AP) of 0.458 and maintained competitive performance with an overall AP of 0.445. Moreover, we also provide results on the CrackTinyNet dataset; our model achieved around a 0.21 increase in performance. The code, model weights, dataset, and our results are available on \href{https://github.com/msaqib17/Road_Damage_Detection}{https://github.com/msaqib17/Road\_Damage\_Detection}.
\end{abstract}

}

\maketitle
\IEEEdisplaynontitleabstractindextext
\IEEEpeerreviewmaketitle

\section{Introduction}
\IEEEPARstart{I}{nfrastructure} and public facilities, such as roads, play an essential role in the country's economy, particularly in modern-day cities. During periods of rapid economic growth in any country, the focus is heavily on building roads, bridges, and roads. However, many of these infrastructures, such as roads, have aged due to multiple factors, for example, rain, weather, vehicles, and the number of years that have passed, causing various road damages~\cite{arya2021deep}.

Road damage seriously impacts drivers' safety, vehicles' value, and efficiency. Moreover, the number of roads inspected in the next few decades will increase drastically; hence, each country requires a sizeable budget for repair, restoration, rehabilitation, and maintenance~\cite{abdullah2018Smartphone}. According to the Federal Highway Administration, the road network in the United States was 4.17 million miles in 2020, an increase from 3.87 million miles in 1990~\cite{carlier_2022}. Moreover, to construct new roads and maintain the existing ones, the US government spends more than 30 billion dollars each year~\cite{2016DamageCost}. In Europe, 50 million people are wounded annually in crashes~\cite{malkoc}. At the same time, the American Automobile Association documents that actual road damages cost drivers in the United States around \$15 billion over five years, averaging \$3 billion annually~\cite{2016DamageCost}. The primary reason for these statistics is the poor condition of the roads.

The expert engineers identify the affected infrastructure, relying on their knowledge, experience, and background. Due to the increase in infrastructure inspection requirements, numerous cities and municipalities utilize a variety of sensors, such as laser scanners, road profilers, multiple cameras, and 3D sensors, to inspect roads around the world. These sensors capture images of road assets, pavements, and longitudinal and transverse views. Concerning the above-mentioned adverse trend in infrastructure road maintenance and management, efficient, reliable, and sophisticated techniques are urgently required for monitoring. A direct approach is the human visual assessment; regardless, it is laborious, expensive, time-consuming, and prone to human error.

To tackle the issue of manual inspection, many researchers have opted for automatic road inspection techniques to study the road conditions, which can be broadly classified into three categories: 1) laser scanning~\cite{li2009real} methods, which provide accurate information about the status of the road but are expensive and require a road closure, 2) vibration-based~\cite{yu2006vibration} methods are restricted to the touched road elements, and 3) vision-based~\cite{zhang2019automated,maeda2018road,zhang2016road} techniques are inexpensive but lack accuracy. Despite the shortcomings of image-based methods, recent progress in vision-based approaches is yielding exceptional results and thus extending their benefit for diverse applications such as road sign detection~\cite{lee2018simultaneous}, traffic analysis~\cite{castaneda2021bim}, crowd counting~\cite{tahir2021transformers}, object detection~\cite{joseph2021towards}, etc.

It should also be noted that specialized vehicles with different sensors, including laser scanners, road profilers, multiple cameras, and 3D sensors, are employed to inspect roads by capturing images of road assets, pavement images, and longitudinal and transverse image profiles~\cite{marcelino2018comprehensive}. Although these vehicles are inexpensive and more efficient than the traditional human visual surveying approach, they can still reach millions of dollars per vehicle, depending on the sensors acquired to build the system to be mounted on the vehicle~\cite{radopoulou2015detection}. Meanwhile, handheld equipment such as smartphones is readily available with high-resolution cameras, robust sensors, and processors. Therefore, vision-based techniques are evolving rapidly and becoming more widespread. We also anticipate that smartphones will play a critical role in controlling the cost of road inspection.

Recently, researchers attempted to utilize vision-based techniques to inspect the road surface by exploiting convolutional neural networks and deep learning methods. For this, the research direction is either on detecting damage on the road surface irrespective of its type or classifying road damage type into a specific class~\cite{zhang2019automated}. The pioneering works focused on detecting directional cracks, i.e., horizontal and vertical~\cite{zhang2016road}, followed by detecting three road damage categories, i.e., horizontal, vertical, and alligator  detection~\cite{akarsu2016fast}. Likewise, the comprehensive classification of the damage types in the road is provided by~\cite{maeda2018road} because distinguishing damage types is crucial for accurate route planning.

Thus far, vision-based inspection approaches lack common ground and suffer from various shortcomings, such as 1) the results being compared on non-standard datasets, where each algorithm uses its own road damage dataset. We aim to create a standard dataset, drawing inspiration from other research areas that have established standard benchmark datasets, such as DnD for denoising, ImageNet for object classification, and PartNet for point clouds. 2) The datasets usually have only one view where the methods are evaluated. 3) The algorithms do not put much effort into the network design for road damage detection or simply employ state-of-the-art, off-the-shelf models from other vision research fields. A specialized network model catering to the needs of road damage detection is warranted. 4) Lastly, road damage can be categorized into many types (e.g., eight damage types in Japan); however, the current research only considers a handful of them; hence, their direct application to practical scenarios is challenging.

\vspace{2mm}
\noindent
\textbf{Our Contributions:} The contributions of this study are as follows

\begin{itemize}
    \item We present a new challenging road damage detection dataset that captures diverse types of damage in individual images under varying conditions, addressing a significant gap in existing datasets. This dataset enables more robust training and evaluation of detection models.
    \item We introduce a 4D Attention mechanism for road damage detection. This enhancement enables the network to process both local details and global context simultaneously, leading to more accurate detection of road damages across different scales.
    \item Through extensive experimental evaluation against state-of-the-art models, we provide comprehensive benchmarking results that show the strengths and limitations of current road damage detection approaches, offering valuable insights for future research directions.
\end{itemize}
\section{Related Works}\label{Related Works}
The maintenance of road infrastructure is essential to ensuring safe and efficient transportation. One of the most significant challenges in road maintenance is detecting and monitoring cracks in the road surface. Cracks in the road surface can cause accidents and lead to costly repairs if not detected and addressed promptly. Machine learning techniques, particularly CNN, are widely applied for automated road crack detection. These techniques have shown promise in improving road crack detection, classification accuracy, and efficiency. However, there has been a limited evaluation of these methods in real-world data and large datasets, which is critical for practical applications. In this section, we analyze recent studies on road crack detection using machine learning techniques, highlighting their objectives, contributions, and limitations. This review aims to provide an overview of state-of-the-art crack detection using machine learning, as well as identify prospective research directions to improve the accuracy and robustness of these approaches.

\subsection{Traditional Methods}
Road damage detection is studied using traditional image processing algorithms, which mainly rely on background subtraction, thresholding, segmentation, and feature extraction. For example, gray-scale histograms with OTSU thresholds~\cite{akagic2018pavement}, gray-level co-occurrence matrix~\cite{sari2019road}, modified median filter and morphological filter~\cite{maode2007pavement}, support vector machine algorithms~\cite{hoang2018artificial}, a library of machine learning models~\cite{gao2020detection}, and edge detectors~\cite{ayenu2008evaluating} are used to detect pavement cracks and different types of road damages. However, these algorithms have limitations and shortcomings, such as sensitivity to lighting and background changes, reliance on handcrafted features, sensitivity to noise, complex backgrounds, and the inability to generalize to various roads and their types, which reduces road damage detection accuracy.

\subsection{Deep Learning Methods}
Recent research~\cite{Zhang2017, Biici2021} has utilized deep convolutional neural networks for automated road assessments and damage identification. Zhang~et~al.~\cite{Zhang2016} trained the supervised deep CNN to classify smartphone pavement images for road damage detection, aiming to find the presence of damage. Similarly, the VGG-based network~\cite{Silva2018} detects cracks on concrete surfaces. Additionally, Crack-pot~\cite{Anand2018} employs a camera-based GPU board to identify road cracks and potholes, which enables smooth journeys for self-driving vehicles and robots. Similarly, Fan~et~al.~\cite{fan2018automatic} trained a simple CNN to recognize pavement conditions, demonstrating its ability to manage various pavement characteristics effectively. Furthermore, the deep neural networks outperform the edge-detection methods in identifying pavement fractures, as shown by~\cite{NhatDuc2018}.

Recently, Karaaslan~et~al.~\cite{Karaaslan2021} analyzed cracks and spalls in a semi-supervised deep learning-based method using attention guidance, where the detected boundary box gets verified by the human, and then the pixel-level segmentation is applied, hence significantly reducing the computational cost of the segmentation. Integrating stereo vision with deep learning, Guan~et~al.~\cite{Guan2021} performed segmentation-based analysis for cracks and potholes by creating datasets composed of 2D, 3D, and enhanced 3D images. Furthermore, for faster segmentation, the authors used a modified U-net based on depth-wise separable convolution and performed 3D image segmentation to measure the potholes' volume automatically. Similarly, the Asfault method~\cite{Souza2018} collected data from a vehicle's accelerometer sensors to record vehicle vibrations and assess road conditions using machine learning.

\subsection{Smartphone-Based Methods}
Due to various sensors, high-resolution multiple cameras, significant storage memory, and effective processors, handheld devices such as smartphones have recently become standard for road inspection. Moreover, handheld devices are efficient and cost-effective for inspecting large road networks. For example, SmartPatrolling~\cite{Singh2017} collects the data employing the smartphone's built-in sensors with dynamic time warping for road surface conditions, performing more efficiently than traditional algorithms. Similarly, Mertz et al.~\cite{mertz2014city} detect road damage using an onboard smartphone installed on daily working vehicles, e.g., general passenger cars, buses, and garbage trucks, and connected to a laptop for processing them. Casas-Avellaneda and Lopez-Parra'~\cite{CasasAvellaneda2016} used smartphones to visualize potholes on a map. Maeda et al.~\cite{Maeda2018} designed a real-time application on a smartphone for collecting and detecting roadway deficiencies.

\subsection{Datasets for Road Damage Detection}
Maeda et al.~\cite{Maeda2018} composed RDD-2018, which is publicly available and is collected through a smartphone application. Several municipalities are using this application to monitor road conditions faster. Consequently, researchers worldwide have shown interest in the data, methods, and models. Following~\cite{Maeda2018}, researchers have either added more images to the dataset or used the software to collect a novel dataset. For example, the authors of~\cite{Angulo2019} expanded the dataset by including images from Italy and Mexico, increasing the number to above 18k. Similarly,~\cite{Roberts2020} incorporated 7k images of road damage from Italy, utilizing the software from~\cite{Maeda2018}. The model is trained on the new dataset, taking into account the identified road damage severity. The authors in~\cite{Du2020} use an industrial high-resolution camera to collect a dataset of over 45k road images from Shanghai and use the YOLO model to detect and classify pavement damages, using an industrial high-resolution camera to compile a dataset exceeding 45k road photos and utilizing the YOLO to detect and classify pavement problems.

On the other hand, Google Street View pictures of pavement, which are easily and freely accessible, are used by~\cite{Majidifard2020}, which include top and wide angles characterizing and calculating the density. The study demonstrates that the Faster R-CNN lags behind the YOLO-v2 model. Moreover,~\cite{Patra2021} employed limited Google API images for CNN-based detection of potholes. Despite the accessibility and cost-free nature of Google Street View photographs, annotating them remains demanding, laborious, and time-intensive. Recently, RDD~\cite{Arya2021} has been proposed for detecting and classifying road damage, which has many images. However, most photos only feature single damage, and the damages themselves are not particularly complex. On the other hand, the damages in our collected dataset are more complicated and typically occur in multiple instances within one image. Table~\ref{tab:DatasetSOTA} shows more information the datasets.

In summary, the studies reviewed demonstrated that machine learning techniques, particularly deep CNNs, accurately detect road cracks. However, the evaluations were limited to small datasets, and there was a limited evaluation of real-world data, which is crucial for practical applications. Additionally, some studies lacked detailed explanations of the proposed CNNs, making it challenging to reproduce the results.

\section{Datasets}\label{Datasets}
We have generated and collected the most challenging road damage dataset. Our dataset called Diverse Road Damage Dataset (DRDD) contains $1500$ images with $5$ damage types using GoPro cameras during different weather conditions and daytimes. The resolution of images in our first dataset is $1920 \times 1440$ pixels. We have verified with researchers and municipality planners that additional views would be convenient as an initial phase. The dataset contains $5$ distinct classes with a real-world long-tailed class imbalance and unique class distribution, as shown in Figure~\ref{ground_truth_analysis}. Moreover, the data annotation project involved 12 people with professional skills and diverse backgrounds, taking a large number of person-hours effort. Furthermore, we have a few rounds of quality assessment for each batch of data.

\subsection{Dataset Characteristics}
\noindent
\textit{Diverse Conditions.} DRDD is collected for diverse scenes and weather conditions. The images contain rich backgrounds such as trees, buildings, architecture, and other vehicles. Different from current datasets, our images in the datasets have more diverse distributions of objects in scenes. The lighting conditions are also different due to different day times, influencing the color of the pixels in the images. Similarly, the images are taken in clear and overcast conditions to capture the variation. The weather conditions affect the damage type and the background, which makes it challenging for many algorithms to distinguish between the weather-induced structures and the actual damage type.

\vspace{2mm}\noindent
\textit{Damage Densities.}
Our dataset contains various density scenarios, divided into three levels: fewer than two, two to three, and more than three damage types. The high densities are one of the most crucial characteristics of our dataset. Table~\ref{tab:DatasetSOTA} compares related datasets widely used for the damage detection types of large-scale outdoor scenes. Our datasets have three notable evaluation parameters: i) the number of damage types per image, our dataset has $3$ damages on average, which is more than the existing datasets, ii) the distribution of the damage types in the scene showing the ratio of road damage, where the statistics show our exceeds others in density, and iii) the degree of damages by computing the mean number of damages in meters centred around each damage type.

\begin{table*}[t]
\centering
\caption{Summary of road image datasets and research works.}
\resizebox{\linewidth}{!}{
\begin{tabular}{l|l|l|l|l|cc|c|cc|cc|ccc}
\hline
     
											&    							&   				&	\textbf{No. of}   				&  \textbf{Image}	            					& \multicolumn{2}{c|}{\textbf{Acquisition}}		& \textbf{Moving}   		&\multicolumn{2}{c|}{\textbf{View}}	& \multicolumn{2}{c|}{\textbf{Damage type}}   & \multicolumn{3}{c}{\textbf{Label type}}\\ \cline{6-15}
\textbf{Dataset}    						& \textbf{Methodology}    		& \textbf{Location}  & \textbf{Images}   & \textbf{Resolution} 	            & \textbf{Camera} 	    & \textbf{Smartphone} 	& \textbf{Camera}    		& \textbf{Top}  & \textbf{Wide}   	& \textbf{Cracks} & \textbf{Potholes} 	& \textbf{Pixel} 	& \textbf{Bbox}         & \textbf{Image} 	\\ \hline \hline
Ouma and Hahn~\cite{Ouma2017} 				& Traditional Algorithms  		& Kenya 			& 75 	    		& 1080$\times$1920 	             	&						& $\checkmark$			& 				    		& $\checkmark$  & 				  	&				  & $\checkmark$ 	  	& $\checkmark$ 		&  						&  					\\ 
CrackIT~\cite{Oliveira2014}    				& Traditional Algorithms  		& - 				& 84 	    		& 1536$\times$2048	             	&$\checkmark$			&  						& - 						& $\checkmark$  & 				  	& $\checkmark$    & 				  	& $\checkmark$ 		& $\checkmark$ 			&  					\\                   
CFD~\cite{Shi2016}		 					&Traditional Algorithms   		& China 			& 118 	    		& 480$\times$320 	             	&						& $\checkmark$ 			&   						& $\checkmark$  & 				  	& $\checkmark$    &  					& $\checkmark$ 		&  						&  					\\ 
CrackTree200~\cite{Zou2012} 				& Traditional Algorithms  		& - 				& 206 	    		& 800$\times$600 	             	&$\checkmark$			& 						& - 						& $\checkmark$  &             	  	& $\checkmark$    &  					& $\checkmark$ 		&  						&  					\\ 
Weng et al.~\cite{Weng2019}					& Traditional Algorithms  		& China 			& 217	    		& 2048$\times$1536 	             	&						& - 					& - 						& $\checkmark$  &                 	& $\checkmark$    &  					& $\checkmark$ 		&  						&  					\\ 
SDNET2018~\cite{Dorafshan2018}				& AlexNet  				  		& USA 				& 230 	    		& 4068$\times$3456 	             	&$\checkmark$			&  						& - 						& $\checkmark$  &    				& $\checkmark$    &  					&  					&  						& $\checkmark$ 		\\ 
Li et al.~\cite{Li1970} 					& Back Propagation NN 	  		& China 			& 400 	    		& - 				             	&$\checkmark$			&  						& - 						& $\checkmark$  &                   & $\checkmark$    &  					&  					&  						& $\checkmark$ 		\\ 
Crack500~\cite{Yang2020} 					& CNN					  		& USA 				& 500 	    		& 2000$\times$1500 	             	&						& $\checkmark$			& - 						& $\checkmark$  &                   & $\checkmark$    &  					& $\checkmark$ 		&  						&  					\\ 
EdmCrack600~\cite{Mei2020}	 				& ConnCrack 			  		& Canada 			& 600 	    		& 1920$\times$1080 	             	&$\checkmark$			&  						&   						&  				& $\checkmark$      & $\checkmark$    &  					& $\checkmark$ 		&  						&  					\\ 
GAPsv1~\cite{Eisenbach2017}					& ASINVOSnet 			  		& Germany 			& 1969 	    		& 1920$\times$1080 	             	&$\checkmark$			&  						& $\checkmark$ 				& $\checkmark$  &   				& $\checkmark$    & $\checkmark$ 		&  					&  						& $\checkmark$ 		\\ 
GAPsv2~\cite{Stricker2019}					& ASINVOSnet, ResNet34 	  		& Germany 			& 2468 	    		& 1920$\times$1080 	             	&$\checkmark$			&  						& $\checkmark$ 				& $\checkmark$  & 					& $\checkmark$    & $\checkmark$ 		&  					& $\checkmark$ 			&  					\\ 
Majidifard et al.~\cite{Majidifard2020} 	& YOLOv2, Faster-RCNN 	  		& USA 				& 7237 	    		& 640$\times$640 	             	&$\checkmark$			&  						& 							& $\checkmark$ 	& $\checkmark$      & $\checkmark$    & $\checkmark$ 		&  					& $\checkmark$ 			&  					\\ 
Maeda et al.~\cite{Maeda2018} 				& SSD Inception-v2, MobileNet   & Japan 			& 9053 	    		& 600$\times$600 	             	&						& $\checkmark$ 			&   						&  				& $\checkmark$      & $\checkmark$    & $\checkmark$ 		&  					& $\checkmark$ 			&  					\\ 
Angulo et al.~\cite{Angulo2019}				& RetinaNet 					& Various 			& 18034     		& 600$\times$600 	             	&						& $\checkmark$ 			&   						&  				& $\checkmark$      & $\checkmark$    & $\checkmark$ 		&  					& $\checkmark$ 			&  					\\ 
RDD2020~\cite{Arya2021} 					& MobileNet  					& Various 			& 26,620 			& 600$\times$600, 720$\times$960 	&						& $\checkmark$ 		& 	 						&   			& $\checkmark$      & $\checkmark$    & $\checkmark$ 		&  					       & $\checkmark$ 	   &  					\\ \hline \hline
Ours  (DRDD)                                & RTMDet with Attention4D        & Oman           &1500             & 1920$\times$1440
& $\checkmark$                 &           &                               &                 & $\checkmark$&  $\checkmark$ & $\checkmark$ & & $\checkmark$&   \\ \hline
\end{tabular}}
\label{tab:DatasetSOTA}
\end{table*}


\begin{figure*}[!t]
    \centering
    \begin{tabular}{c@{ }c@{ }c}
   
            \includegraphics[width=0.31\textwidth]{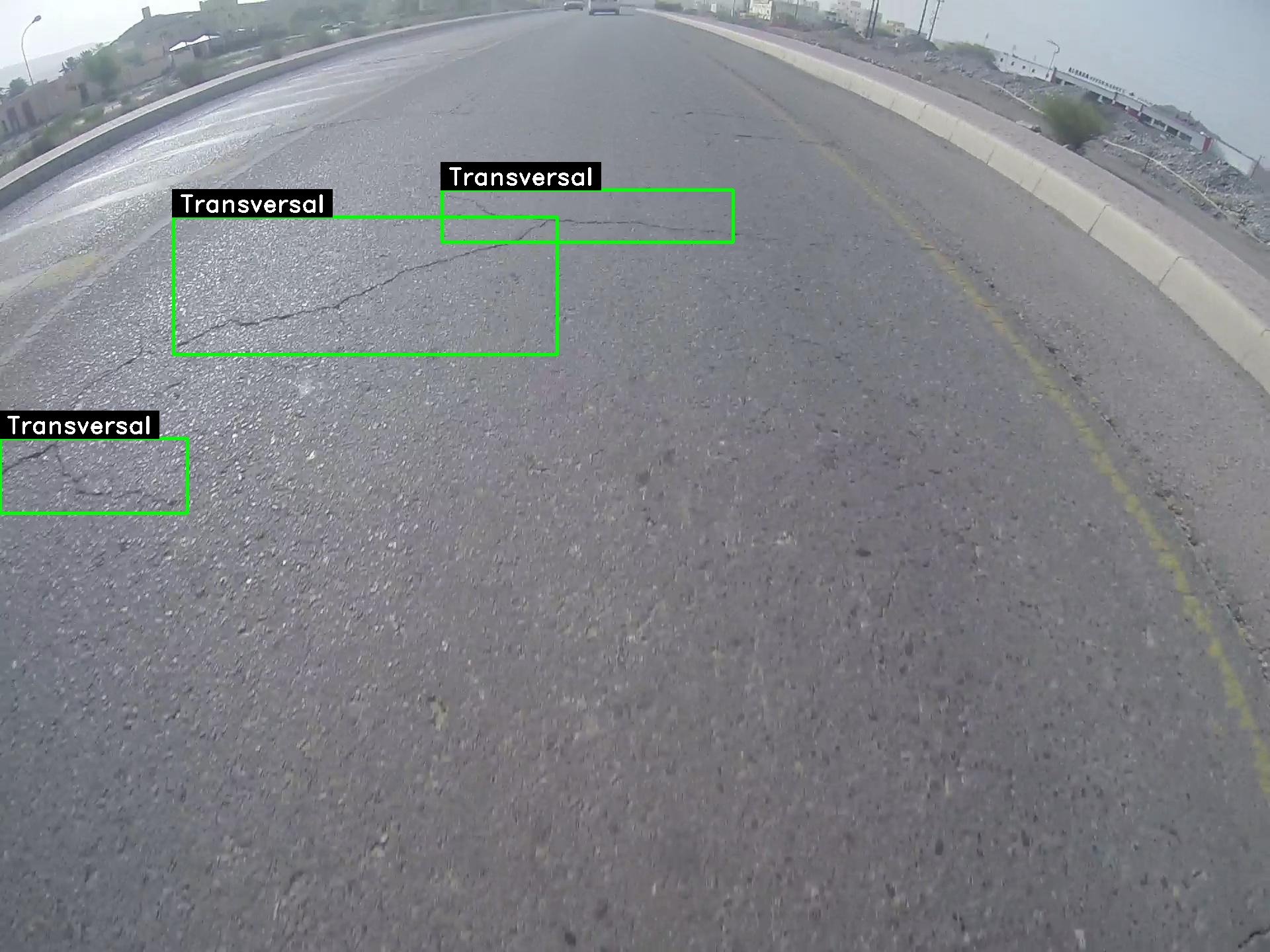}&
            \includegraphics[width=0.31\textwidth]{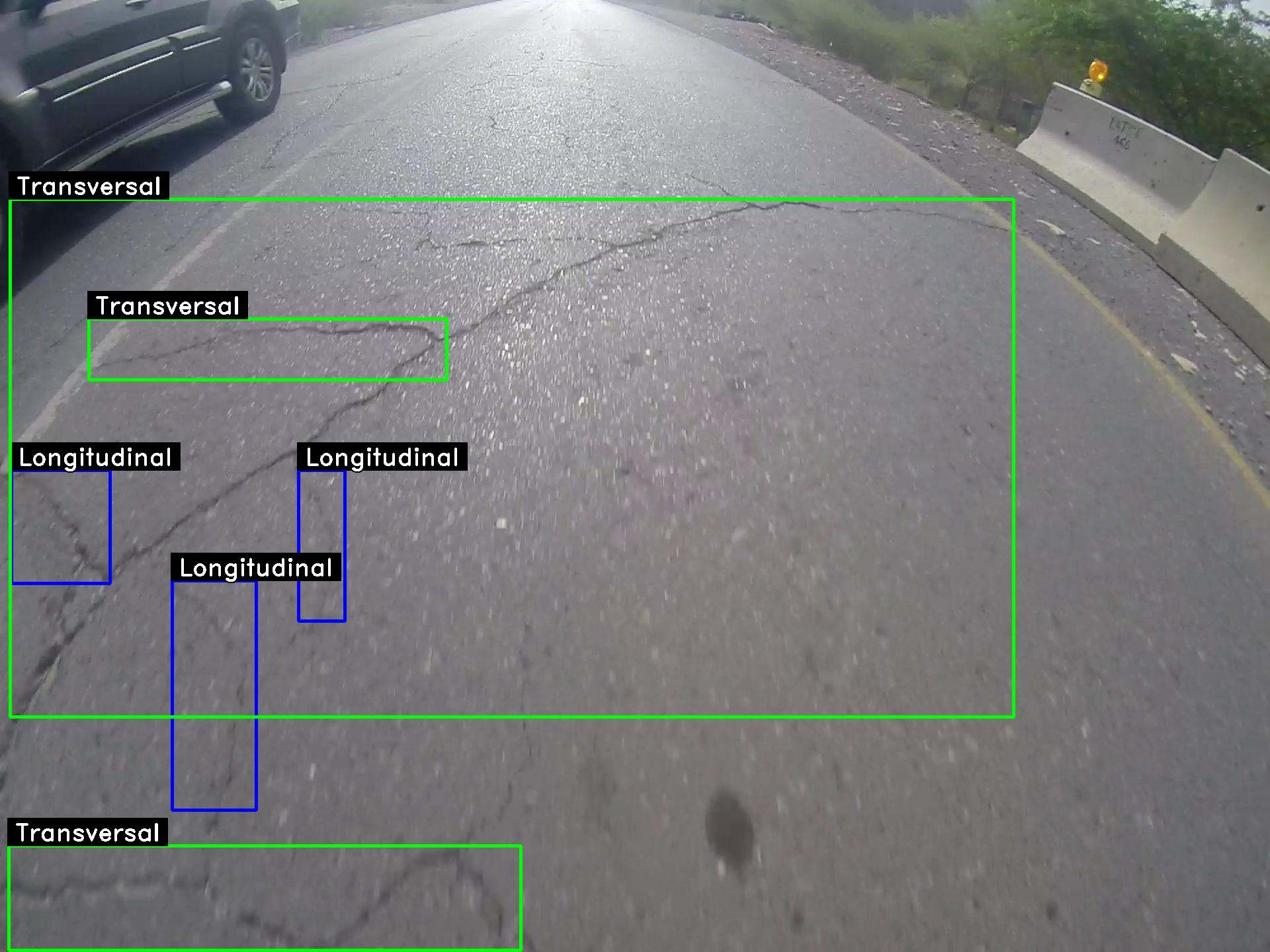}&
            \includegraphics[width=0.31\textwidth]{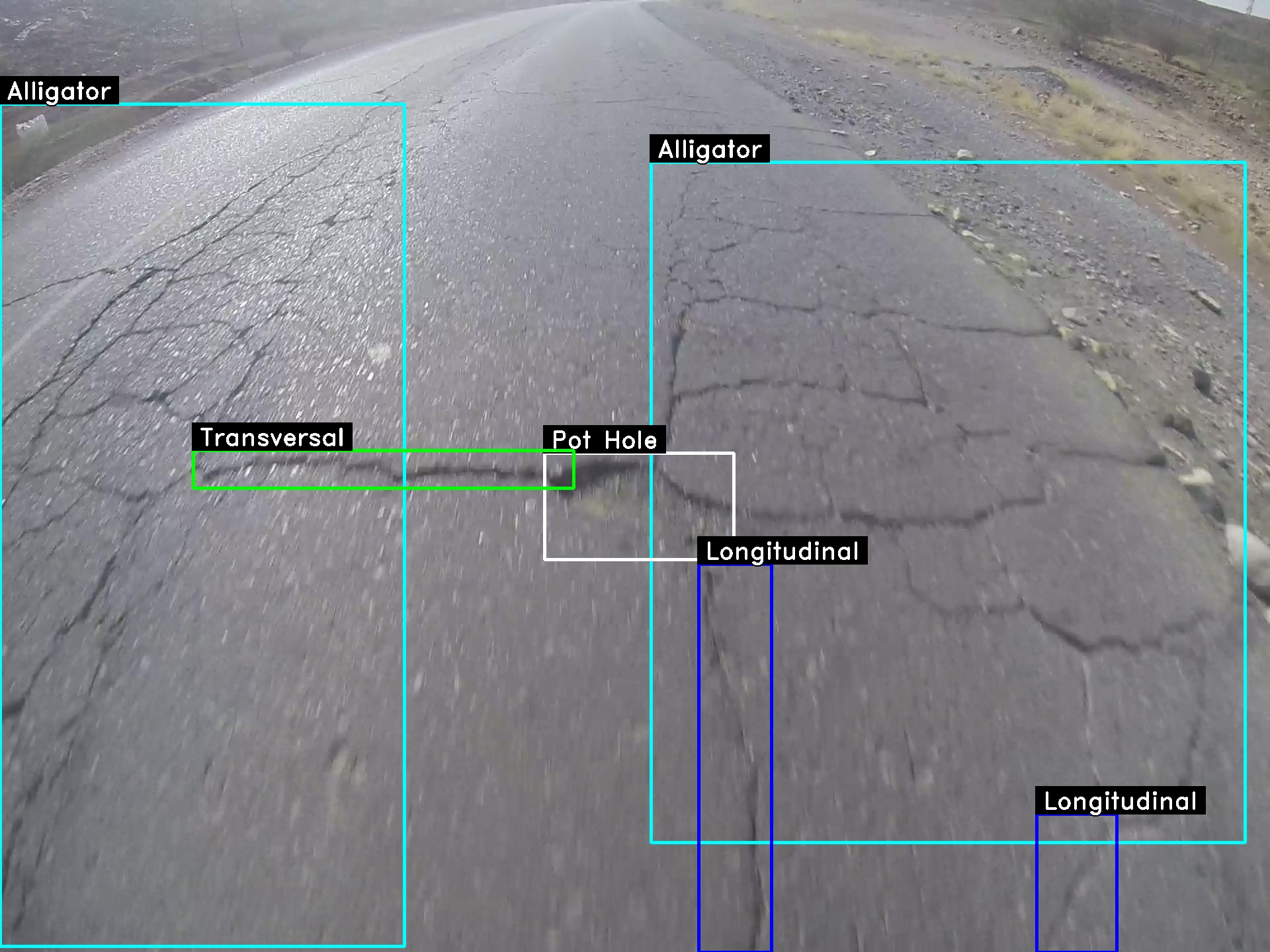}\\
            a) Weather conditions &
            b) Occlusions &
            c) Diverse damage types\\
  
            \includegraphics[width=0.31\textwidth]{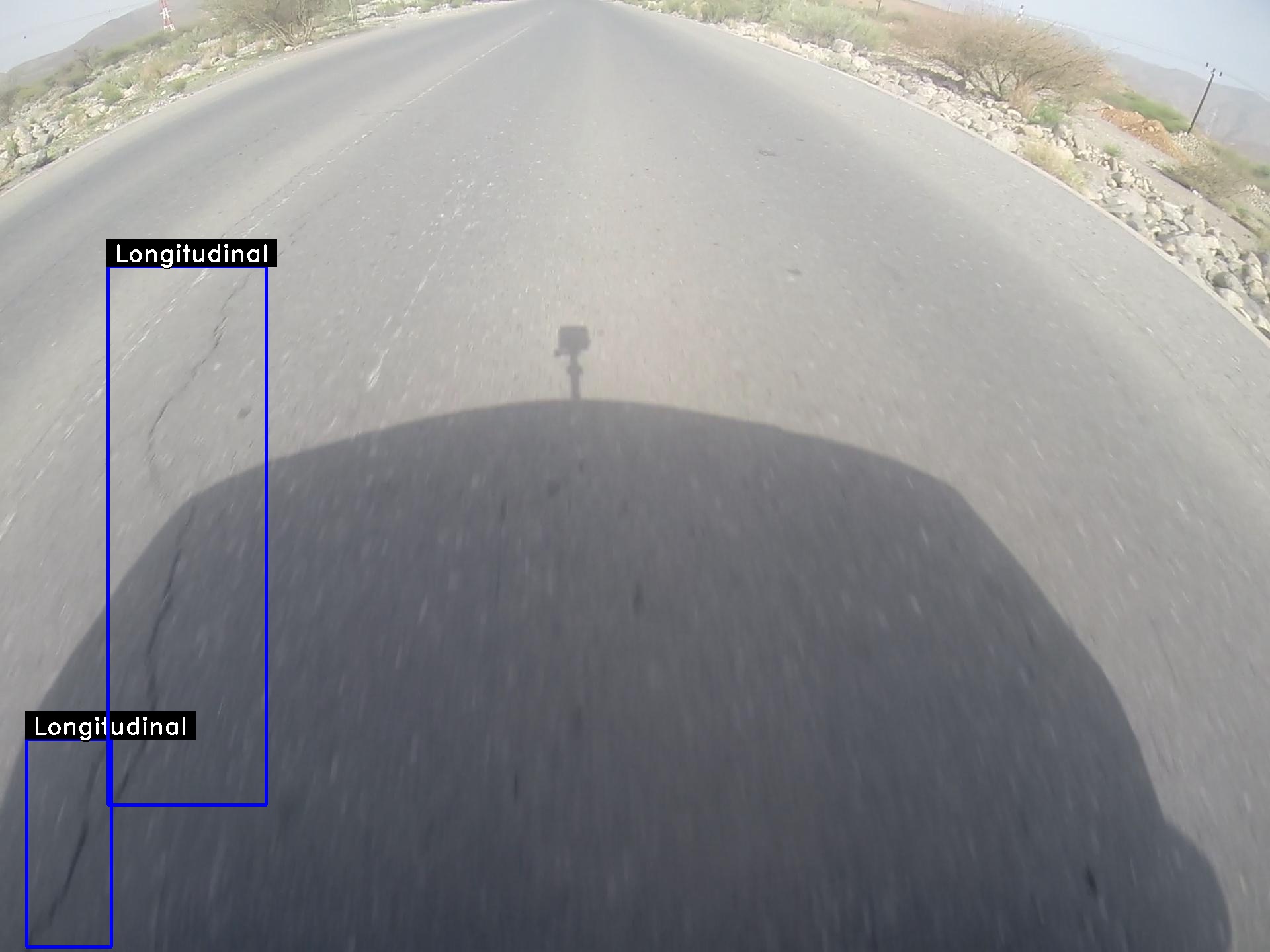}&
            \includegraphics[width=0.31\textwidth]{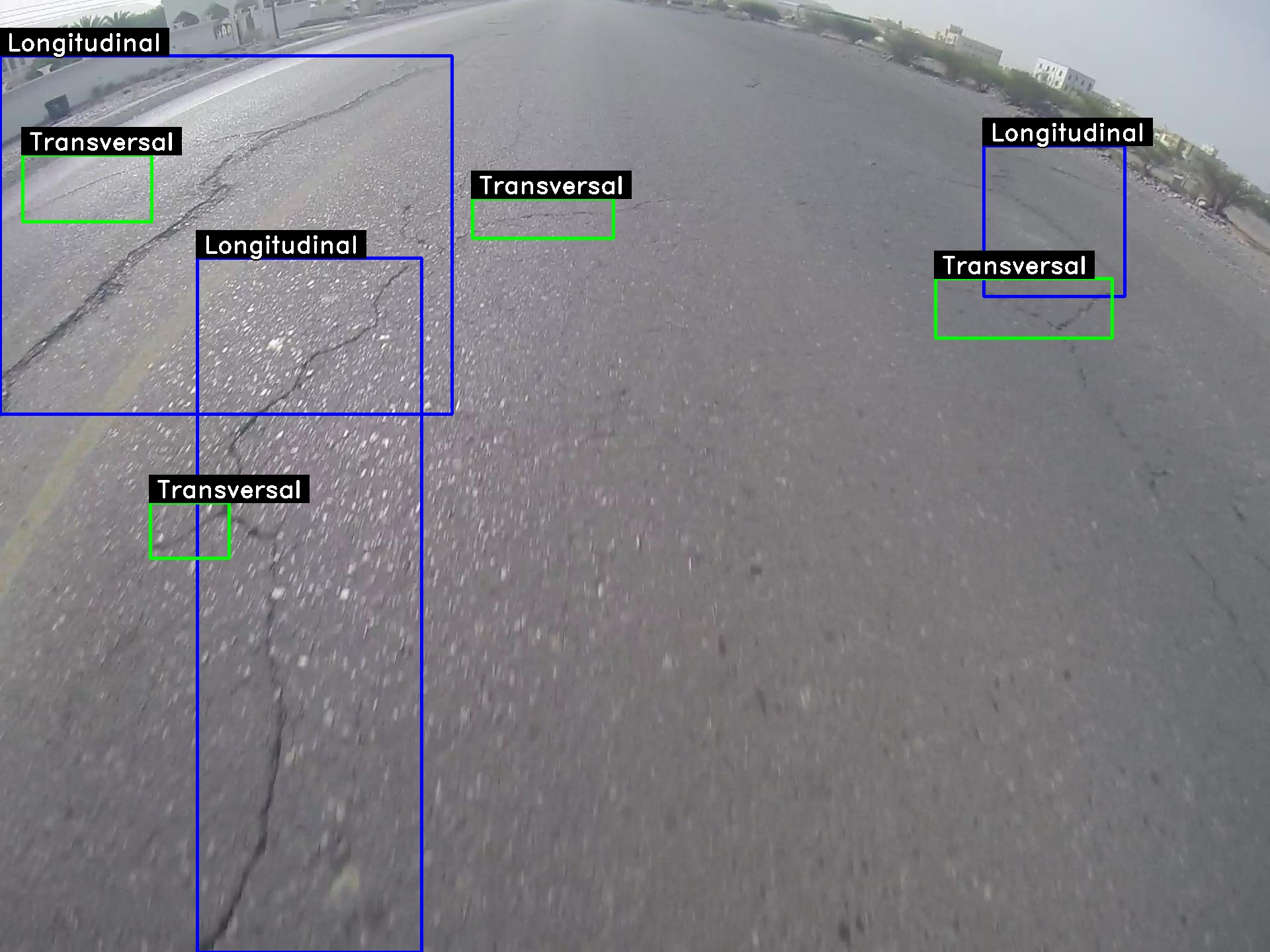}&
            \includegraphics[width=0.31\textwidth]{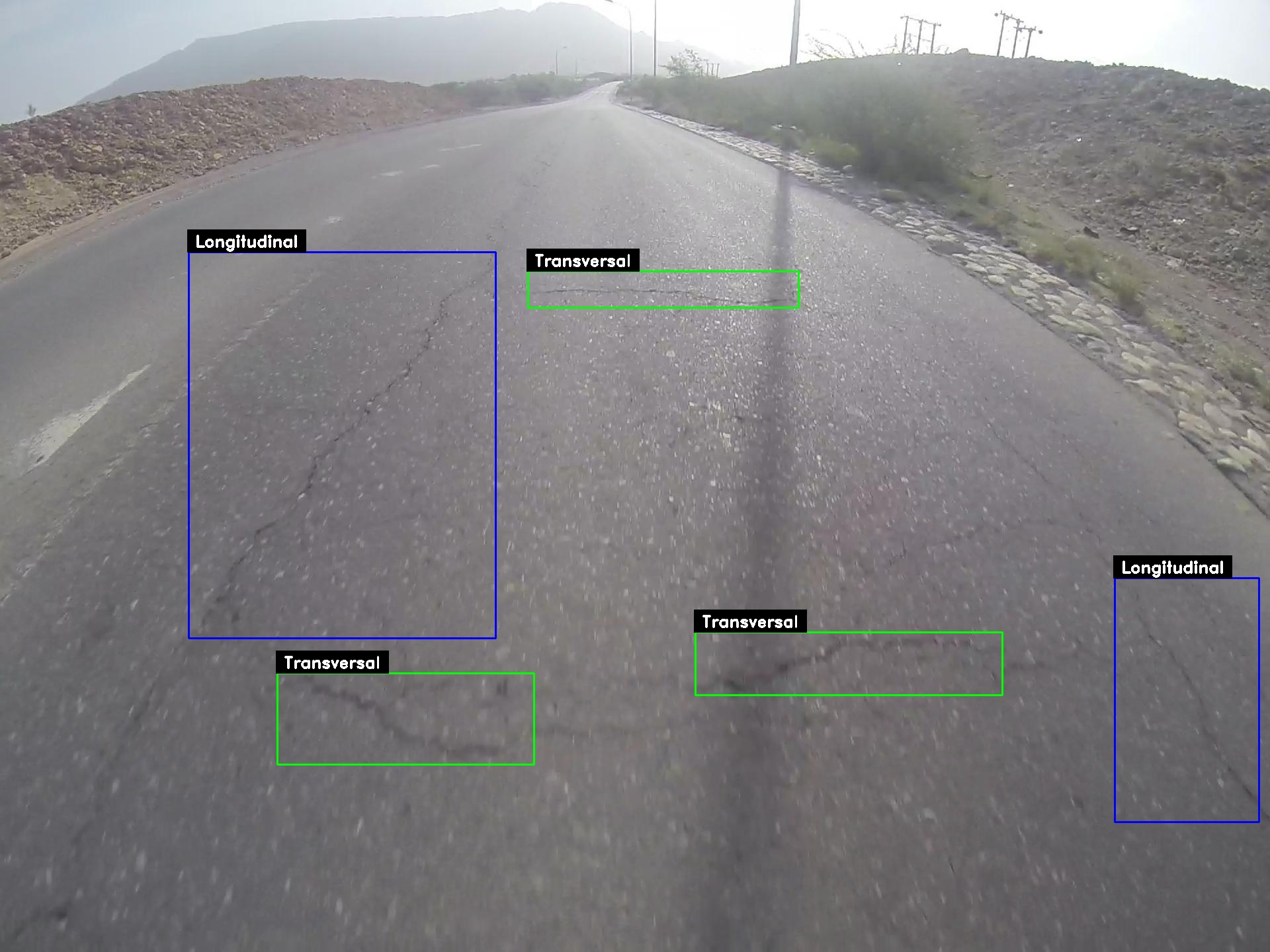}\\

            d) Motion blur &
            e) Instance variations&
            f) Shadow effect\\
   
    \end{tabular}
    \vspace{0.3cm}
    \caption{\textbf{Major challenges in road damage detection:} Representative examples showing various challenges: (a) weather conditions affecting damage visibility and appearance, (b) occlusions from vehicles partially hiding damage areas, (c) diverse damage patterns requiring robust detection capabilities, (d) motion blur from vehicle movement impacting image quality, and (e) instance-level variations in damage characteristics and (d) shadow effect adding complexity to detection.}
    \label{fig:challenges}
\end{figure*}

\vspace{2mm}\noindent
\textit{Occlusions.} DRDD is collected during regular traffic hours to ensure they reflect the actual scenes. However, the one downside is that the dataset's images have occlusions. Figure~\ref{fig:challenges}(b) shows an example of such a scenario where the damage type is partially occluded, making the detection challenging due to limited visible features. According to occlusion situations, the datasets can test the techniques' performance in challenging cases.

\vspace{2mm}\noindent
\textit{Instance-level Diversities.} We show the diverse instance-level densities in Figure~\ref{fig:challenges}(e). The inter-class difference between the damage types is significant. Due to the variation in shape and scale details, long-distance instances are difficult to learn, as damage types differ in size, structure, and appearance, making our datasets challenging for most existing algorithms. 

\vspace{2mm}\noindent
\textit{Diverse Damages Types.} The damage in the roads does not follow a specific pattern; hence, it is possible that a single scene may contain more than one type of damage. This is also reflected in our datasets, where more than one damage type is present. Such images are shown in Figure~\ref{fig:challenges}(c)

\vspace{2mm}
\noindent
\subsection{Dataset Challenges}
We have accomplished a scale much more extensive than any earlier datasets; the scale is essential for generalization, the ultimate goal of any dataset. However, employing the data across cities and countries also raises several challenges, such as noise sources and ambiguities. We manage several known noise sources and ambiguities in our datasets, illustrated in Figure~\ref{fig:challenges}.

\vspace{2mm}
\noindent
\textit{Labeling Noise.} The accuracy of the labels is heavily dependent on the labeller, an expert and a domain specialist; then, it is quite probable that the label will be accurately compared to someone who is a volunteer and learning the process. Moreover, there are no agreed-upon definitions for damage types.

\vspace{2mm}
\noindent
\textit{Image Quality.} The quality and resolution of the images acquired through various cameras and hand-held devices vary significantly. Occasionally vehicles occlude the damage of the road from view, which we protect against by removing images manually, but the possibility of partial occlusion always exists. Furthermore, we also capture multiple views in our datasets to mitigate the occlusion issues.

\vspace{2mm}
\noindent
\textit{Unlabeled Artifacts.} Some artifacts are not damaged to the road but have some temporary or superficial marks, such as drags of the tyres during braking or peel of paint on the road or parts of unwanted items like a branch, trash, etc. These may appear similar to the road damage types; nevertheless, the damage of interest is often the most prominent, and the presence of others can create confusion in detection and localization, which can misguide algorithms.

\vspace{2mm}
\noindent
\textit{Patching Noise.} Road damages are usually fixed using manual techniques, which peels off soon due to temporary materials. Such patching typically hides the structure of the actual damage, creating confusion for the labeler and leading to a discrepancy in the accuracy.

\vspace{2mm}
\noindent
\textit{Unwanted Shadows.} The sun's direction affects the images. If the sun is at the back of the vehicle taking pictures, the vehicle shadow may obscure part of the image, and it is possible that some sections of the damage type as well~\ref{fig:challenges}(d). Hence, it becomes challenging for the algorithms as some regions of the damage type are concealed by the shadow while others are under direct sunlight, causing a difference in the pixels' color.

\subsection{Evaluation Protocol}  
We have established an evaluation protocol to train/test explicitly splits our datasets expressed as follows uniquely.

\vspace{2mm}
\noindent
\textit{Per-Dataset Splits.} 
As per the traditional protocol, we are interested in how well an algorithm generalizes where the train and test splits belong to the same dataset. The training and testing sets are disjoint; here, the training is performed on a defined number of images and testing on another set of images from the same dataset to avoid overfitting on a specific category.



\subsection{Annotations}
We manually labelled high-quality ground truth for the images of both our datasets. We annotated the road damage type using a 2D bounding box (x, y, w, h), where x, y denotes the center coordinates and w, h is the width, and height along the x-axis, and y-axis, respectively. A file is provided for each damage type with the same ID as the image. 

\subsection{Types of Road cracks}
As a last step, we would like to introduce the types of damage in the road. The damage types can be categorized into Alligator Cracks, Block Cracks, Longitudinal Cracks, Pothole Cracks, and Transverse Cracks. These are the most prominent ones and are usually found on most roads across the globe. Figure~\ref{fig:challenges} shows these mentioned cracks.

\subsubsection{Potholes} 
Potholes are holes in the road surface caused by water seeping into the road and eroding the ground underneath. Moreover, the heavy traffic and worn-out surface layer exacerbate the damage by causing the asphalt to break away, forming a pothole. Initially small, these potholes can rapidly expand due to ongoing vehicle traffic that erodes the asphalt further and water from rain or floods that washes away additional material. Proper maintenance and repair of road cracks can help prolong the life of a road and improve safety for drivers. Repair methods can include sealing the cracks, patching the affected areas, and, in more severe cases, repaving the entire road surface.

\subsubsection{Longitudinal Cracks}
These cracks run along the length of the road and can be caused by expansion and contraction of the ground, poor drainage, 
or the settling of the subgrade. They are often caused by the natural movement of the ground and can be found on both the shoulder and the centerline of the road. 
Longitudinal cracks can also be caused by shrinkage of the asphalt surface due to exposure to the sun and heat.

\subsubsection{Transverse Cracks}
These cracks run across the road and are typically caused by temperature changes and heavy traffic. They occur when the road expands and contracts with changes in temperature, causing the asphalt surface to crack. These types of cracks are often found in the wheel path and are typically caused by reflective cracking in the underlying layers

\subsubsection{Alligator or Fatigue Cracks}
These are a series of interconnected cracks that resemble the skin of an alligator. They are caused by a failure in the asphalt surface and heavy traffic loads. They occur when the asphalt surface can no longer support the weight of the traffic and begins to fatigue and deform. These cracks are also known as "fatigue cracking" and can be found on both the shoulder and the wheel path.

\subsubsection{Block Cracks}
These are large rectangle cracks that break the surface into chunks or blocks. It forms when the asphalt surface is too thick and the top layer contracts at a different rate than the bottom, causing the surface to crack.
\begin{figure*}[tbp]
\begin{center}
\begin{tabular}{c@{ } c@{ } c@{ } c@{ } c}

  \includegraphics[width=.19\textwidth]{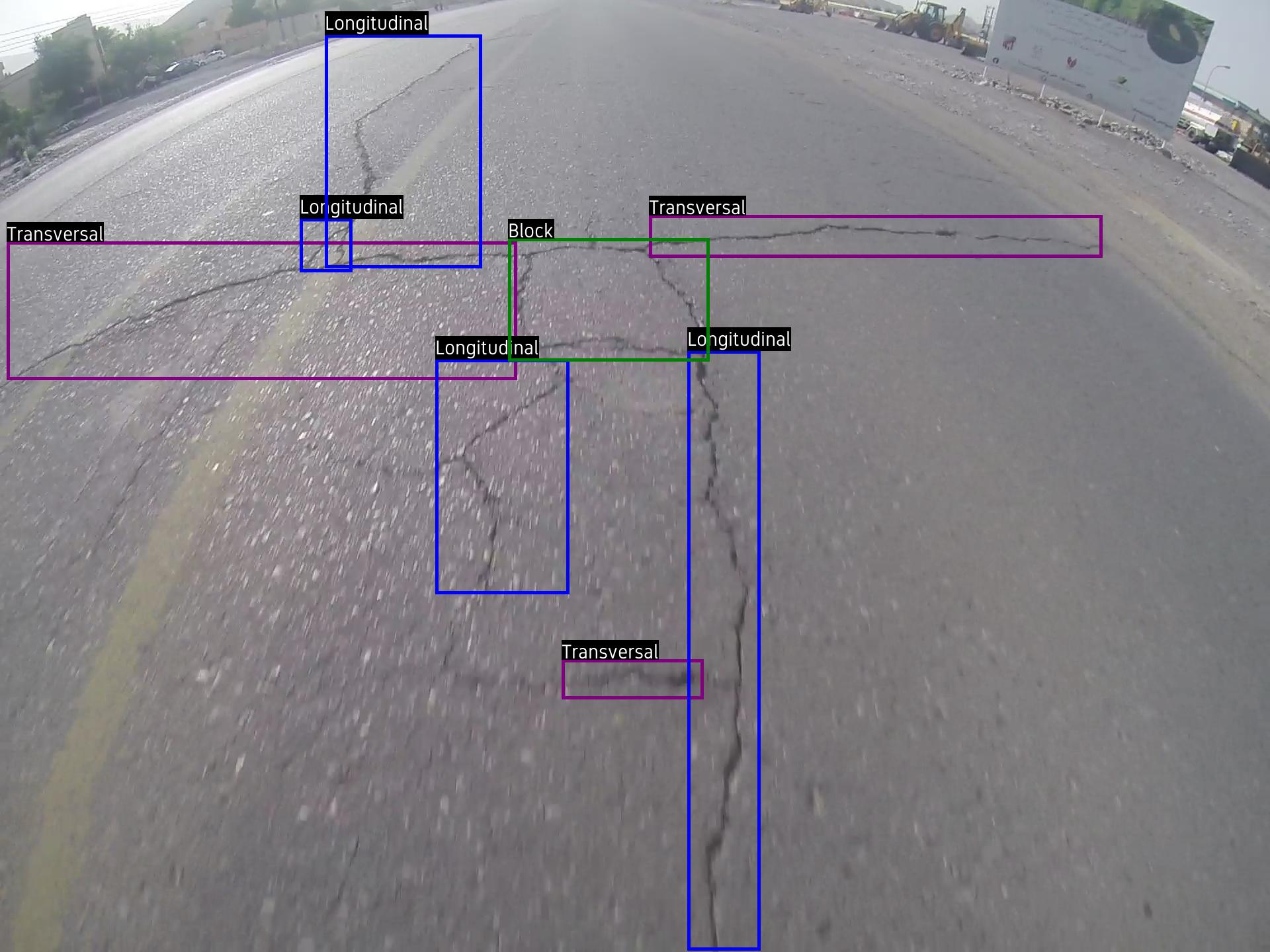}&
  \includegraphics[width=.19\textwidth]{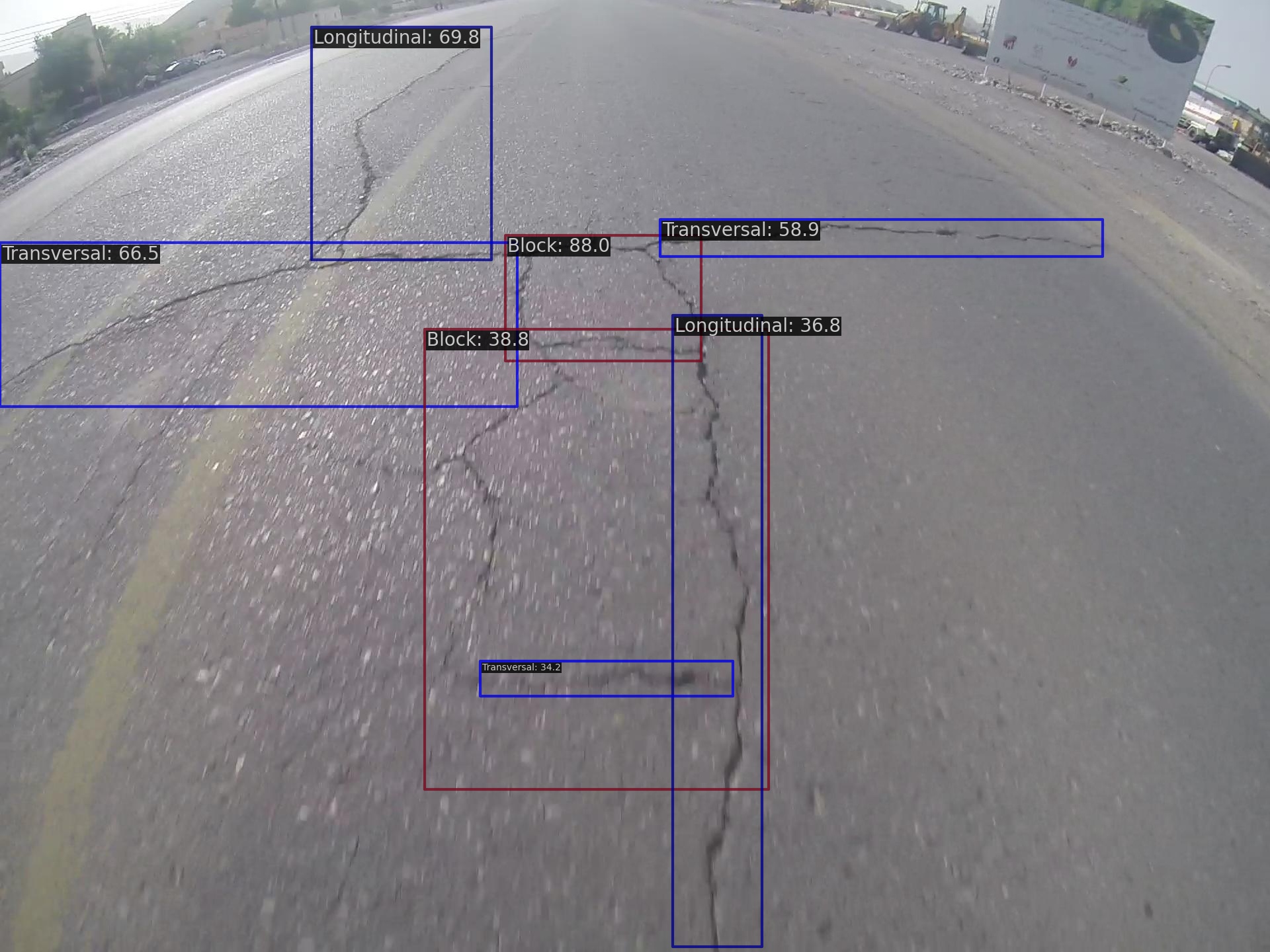}& 
  \includegraphics[width=.19\textwidth]{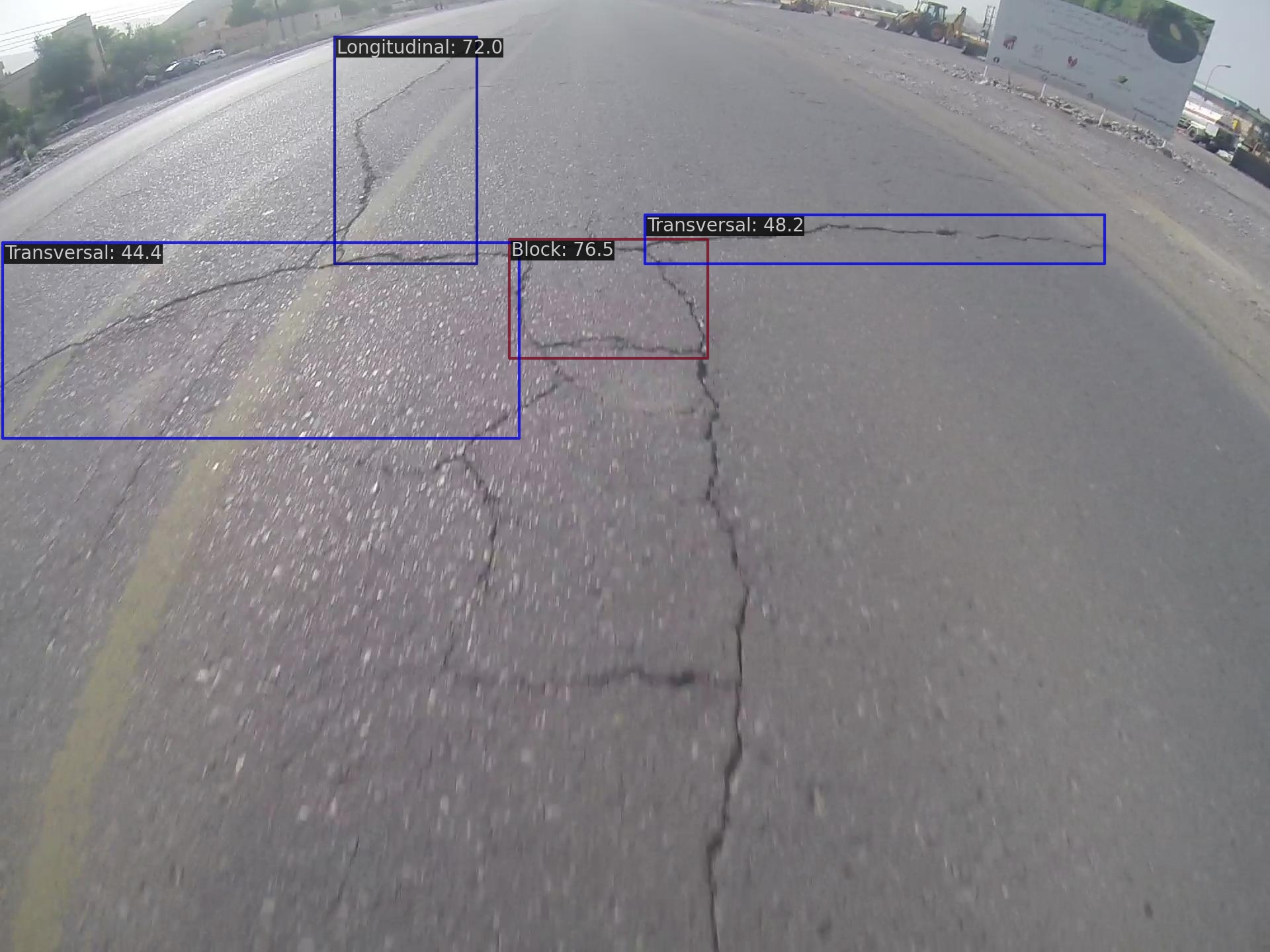}& 
  \includegraphics[width=.19\textwidth]{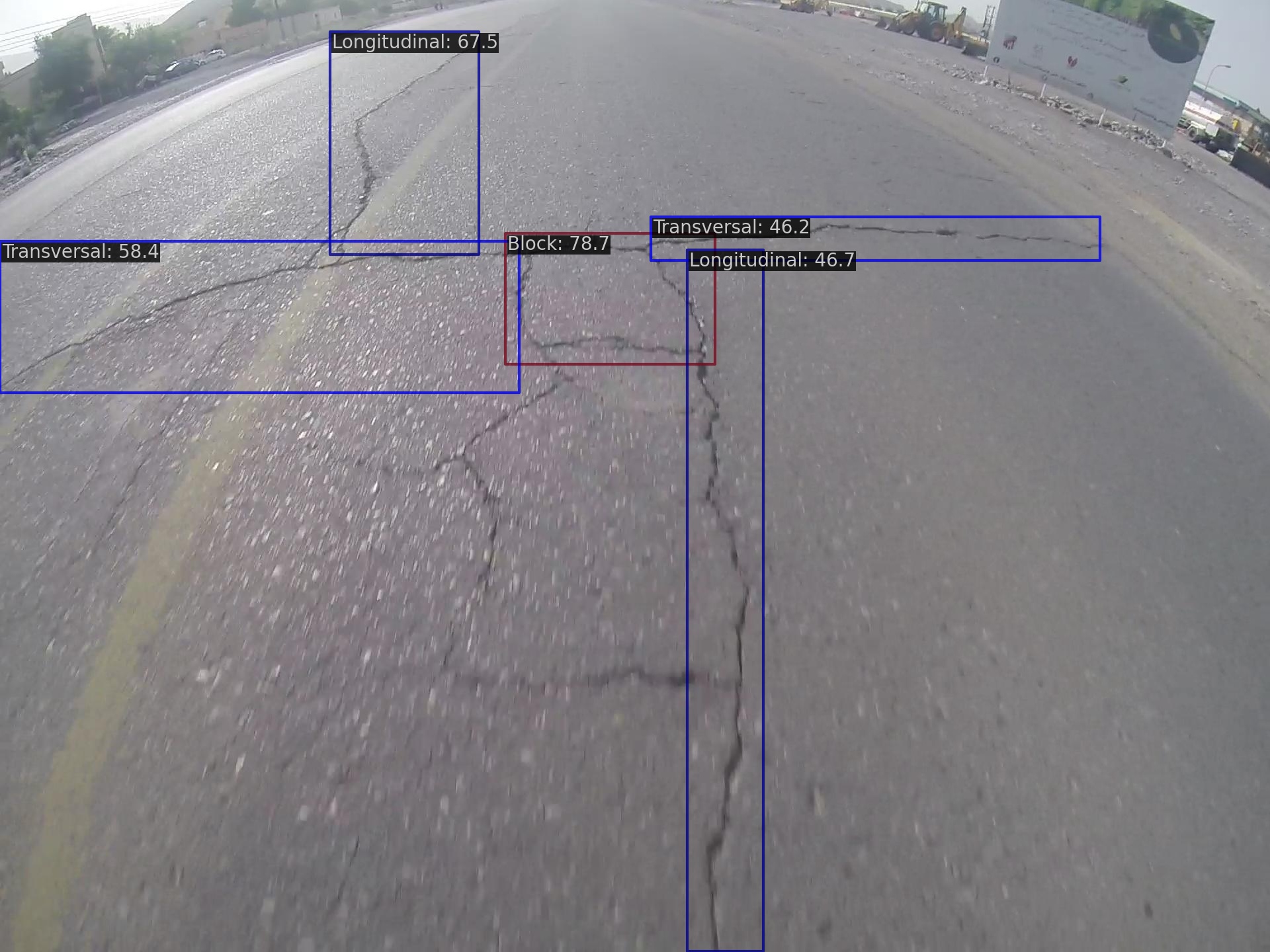}& 
   \includegraphics[width=.19\textwidth]{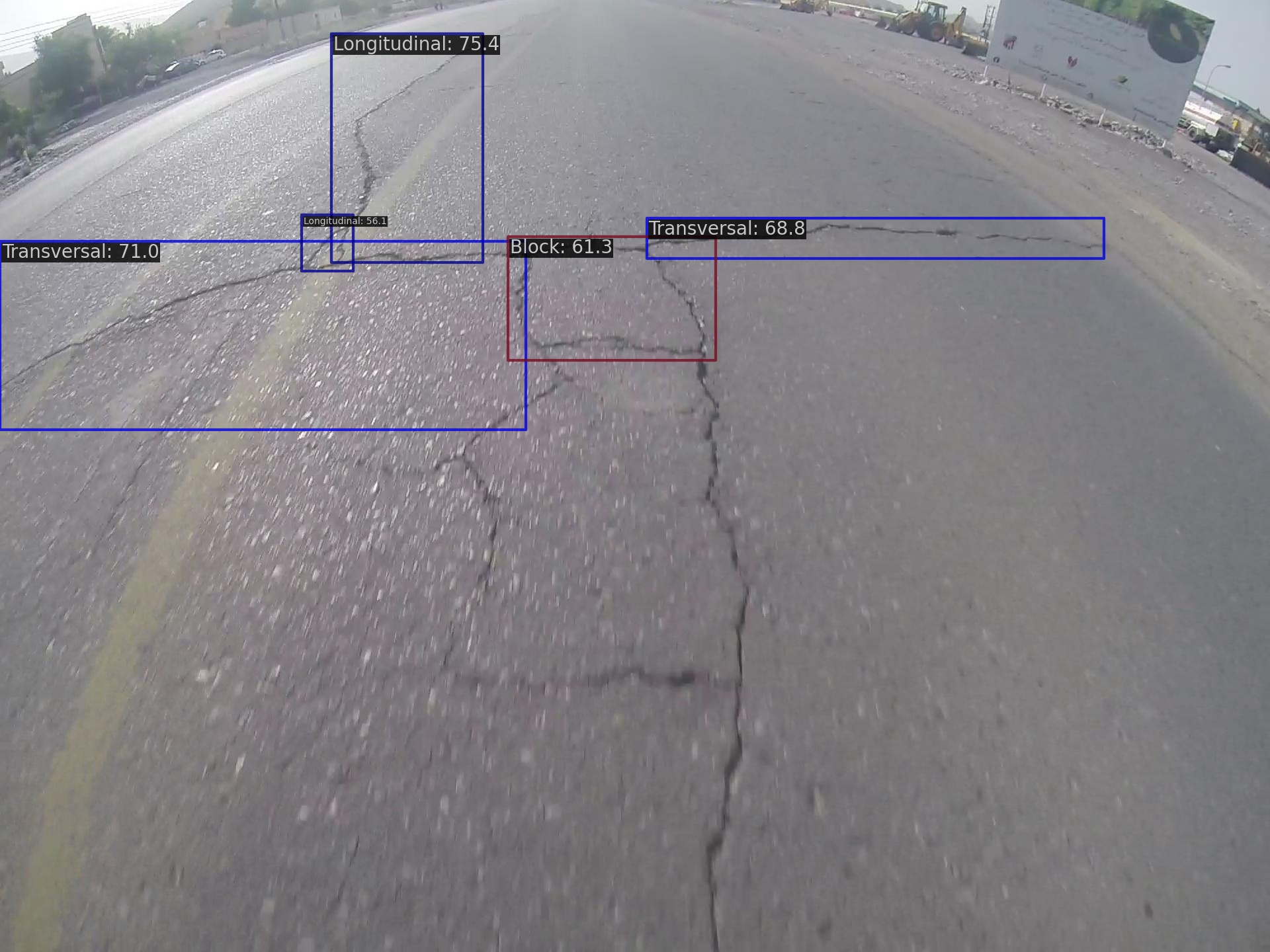}\\

  \includegraphics[width=.19\textwidth]{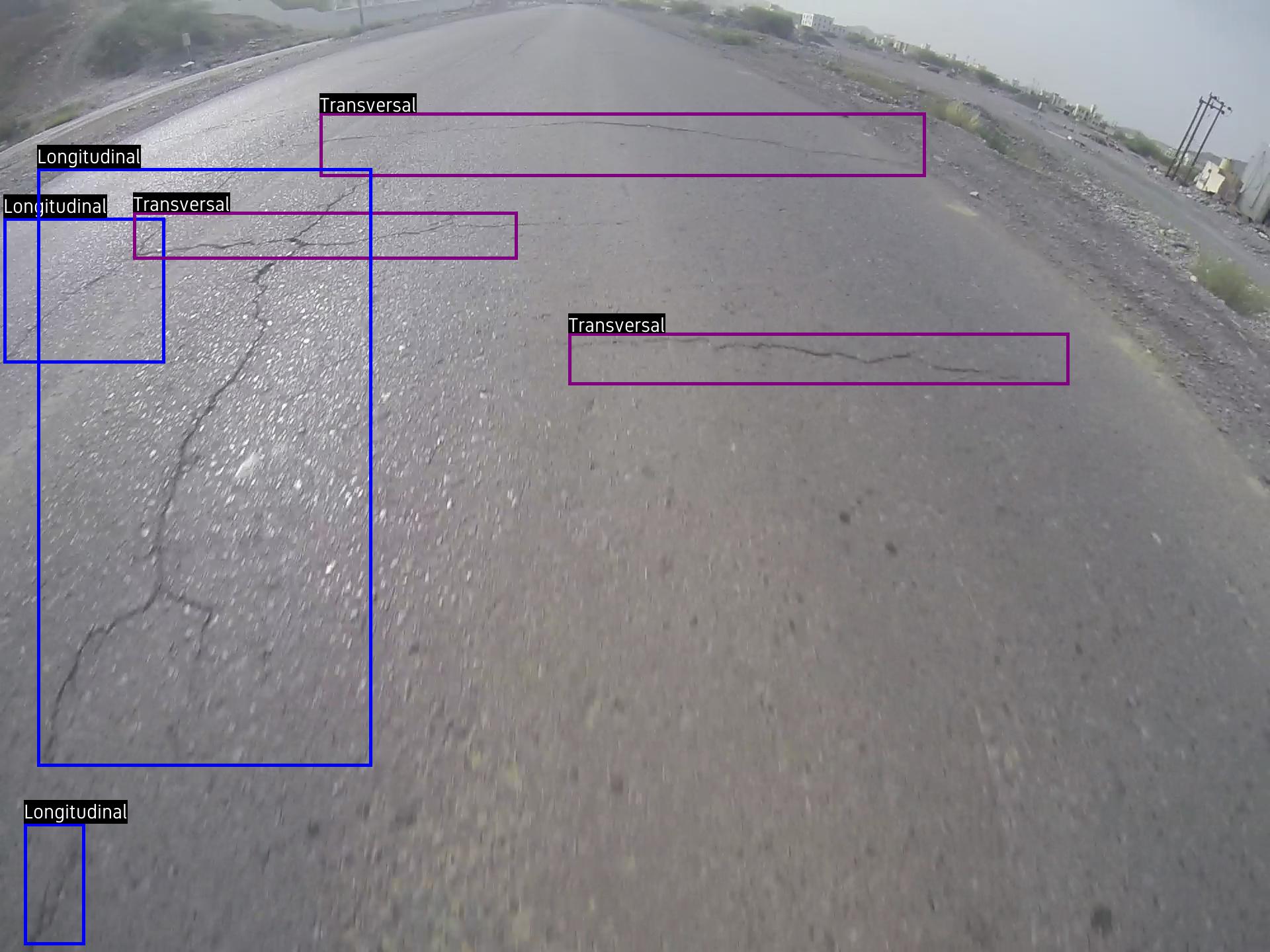}&
       \includegraphics[width=.19\textwidth]{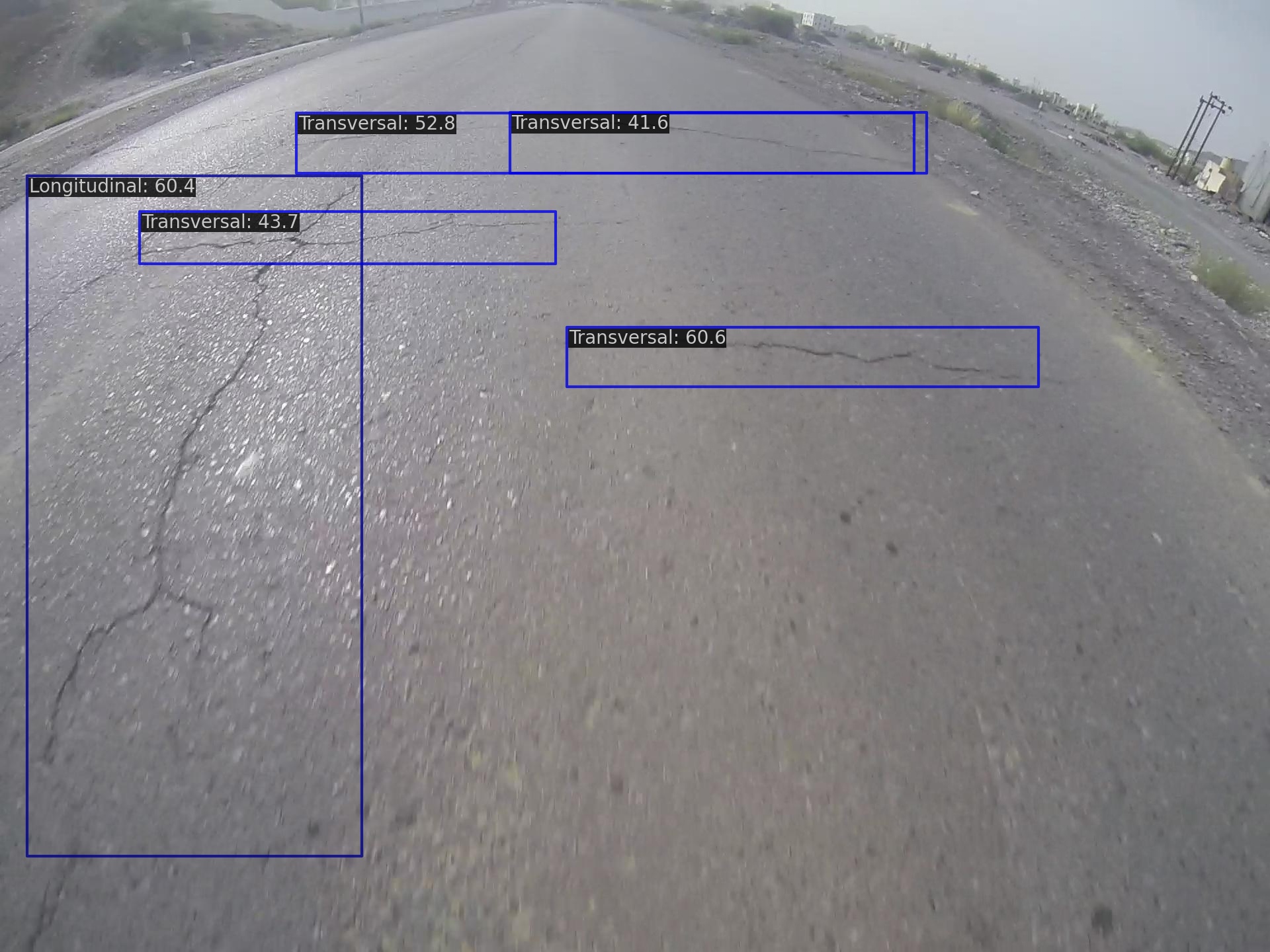}&  
  \includegraphics[width=.19\textwidth]{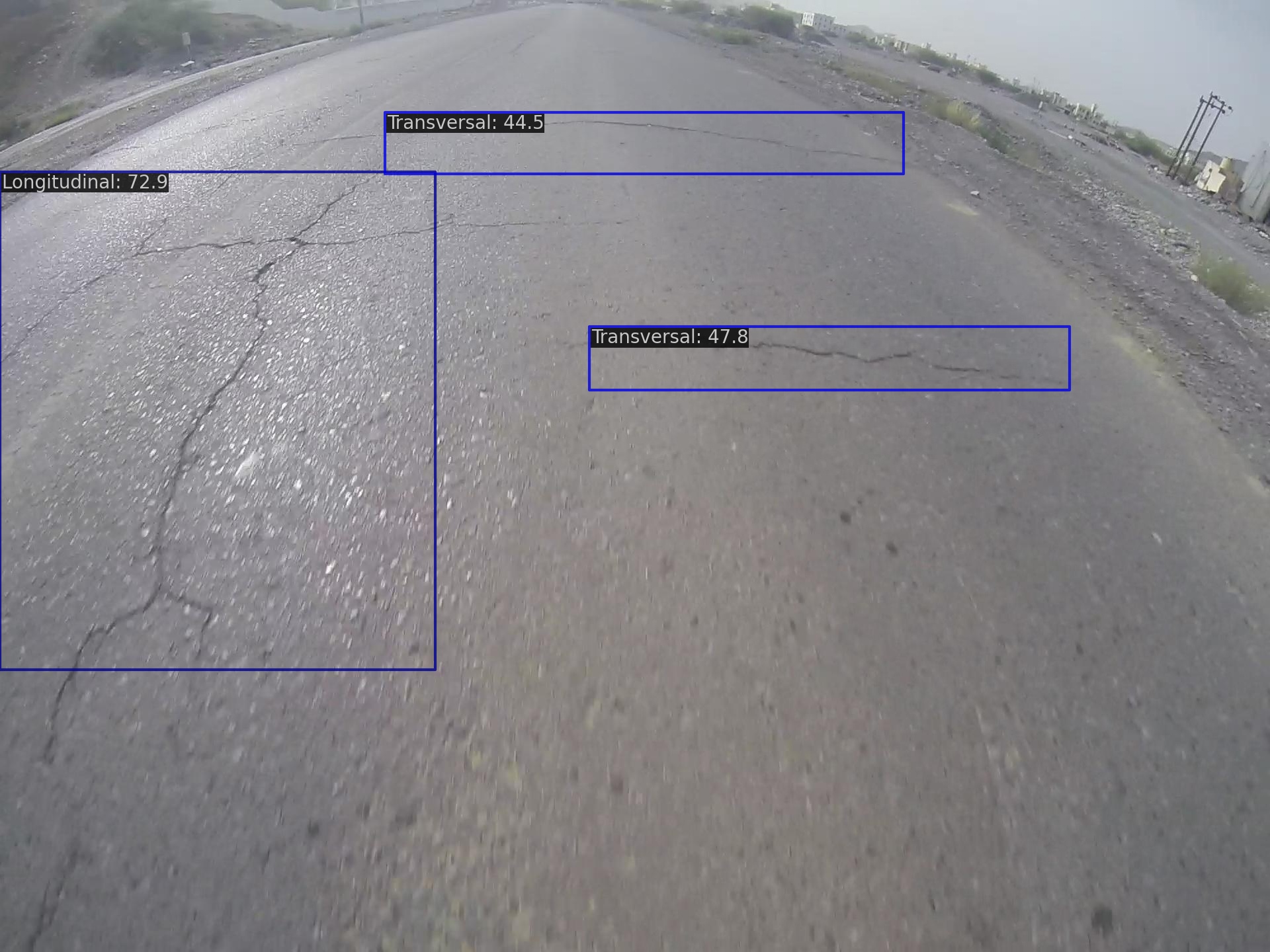}& 
  \includegraphics[width=.19\textwidth]{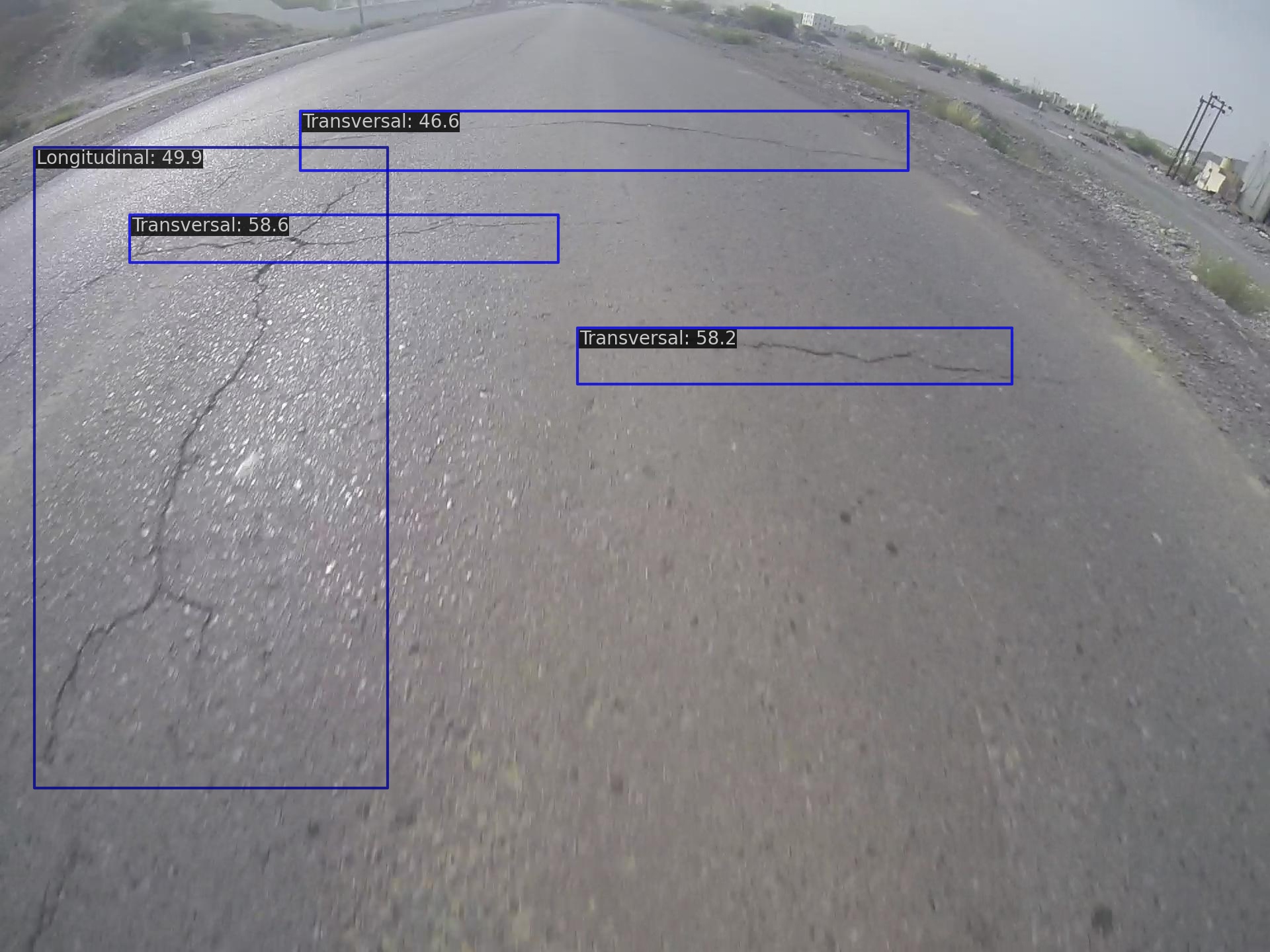}& 
    \includegraphics[width=.19\textwidth]{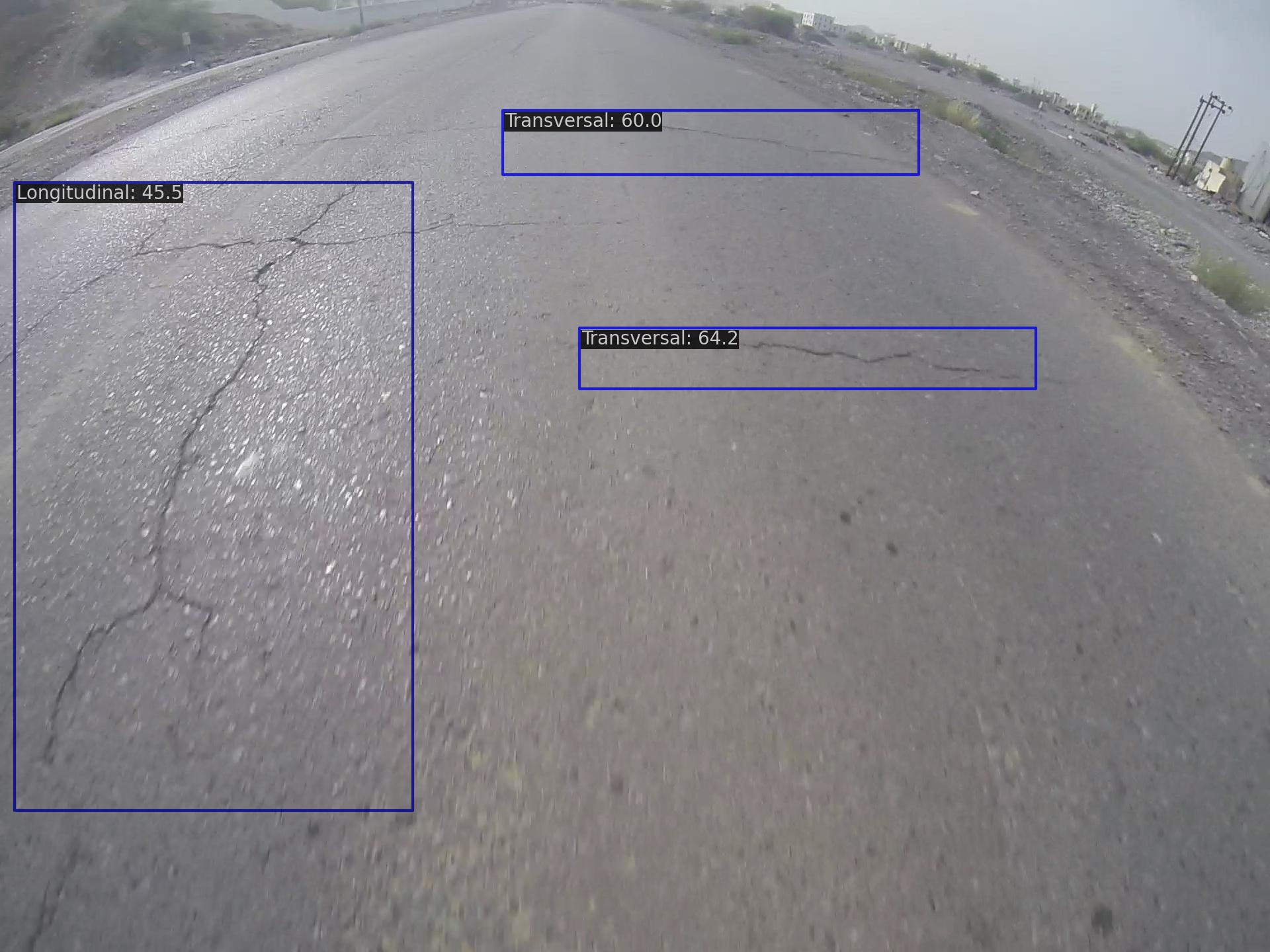}\\

  \includegraphics[width=.19\textwidth]{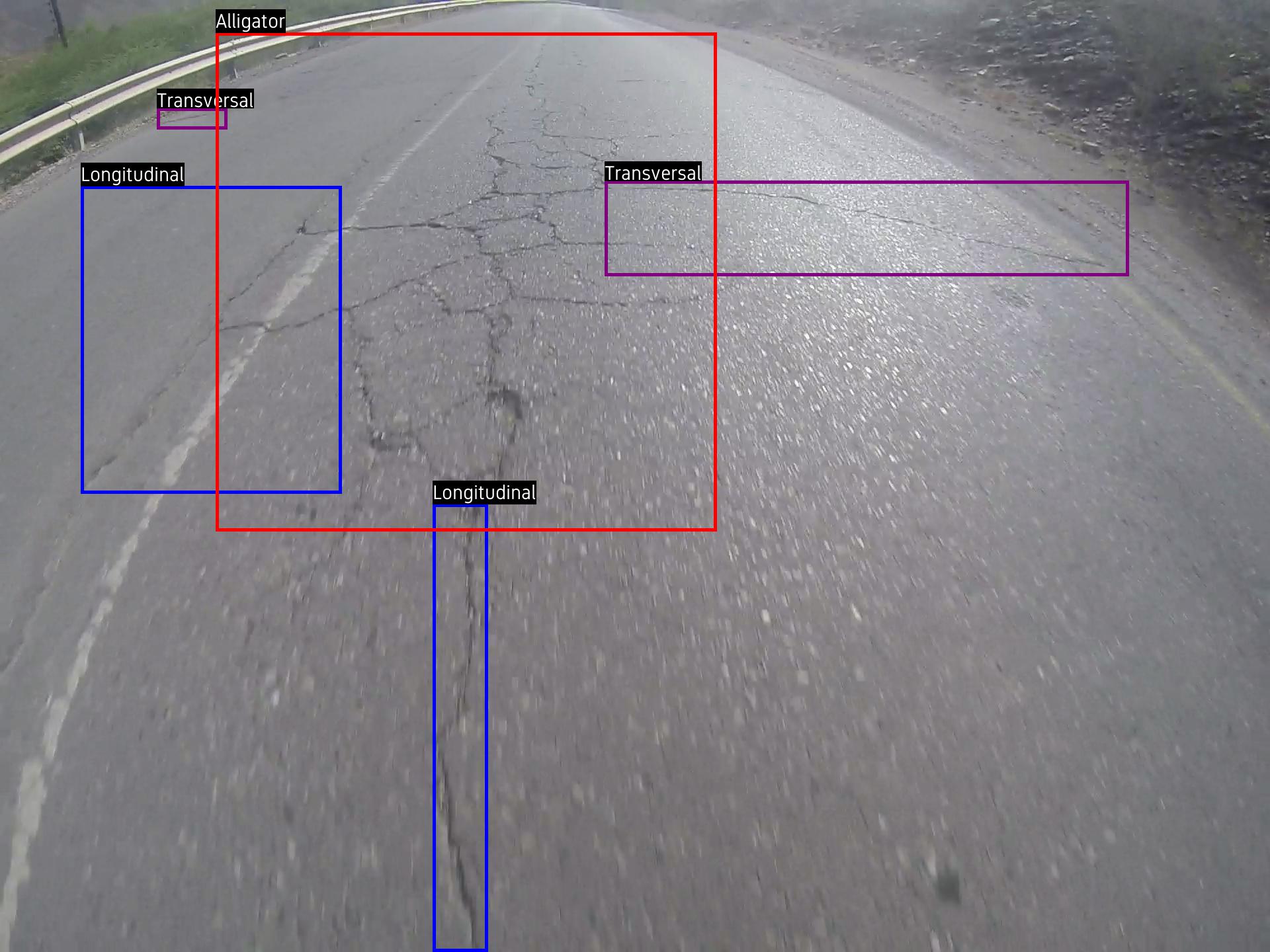}&
  \includegraphics[width=.19\textwidth]{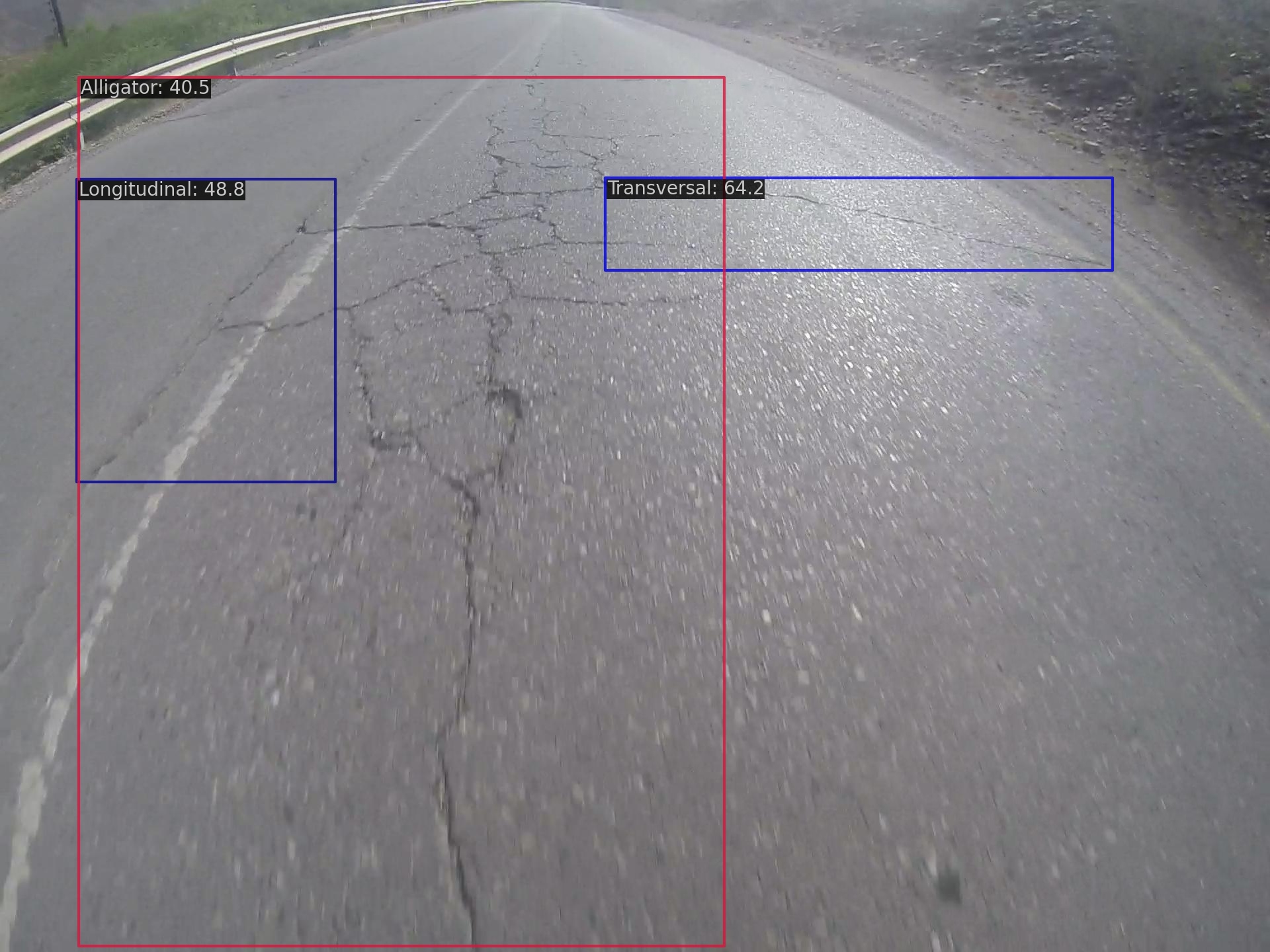}&
  \includegraphics[width=.19\textwidth]{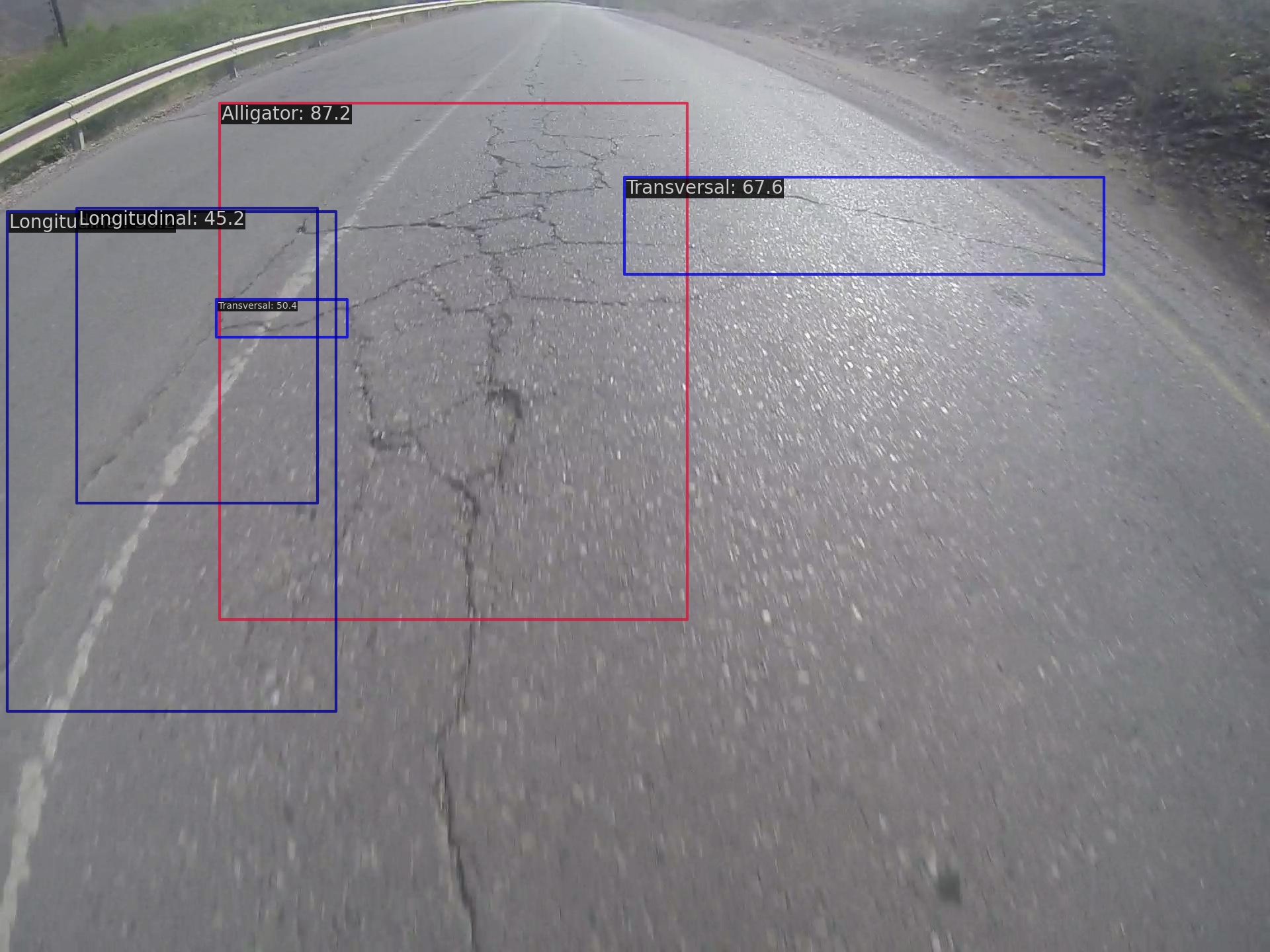}&
  \includegraphics[width=.19\textwidth]{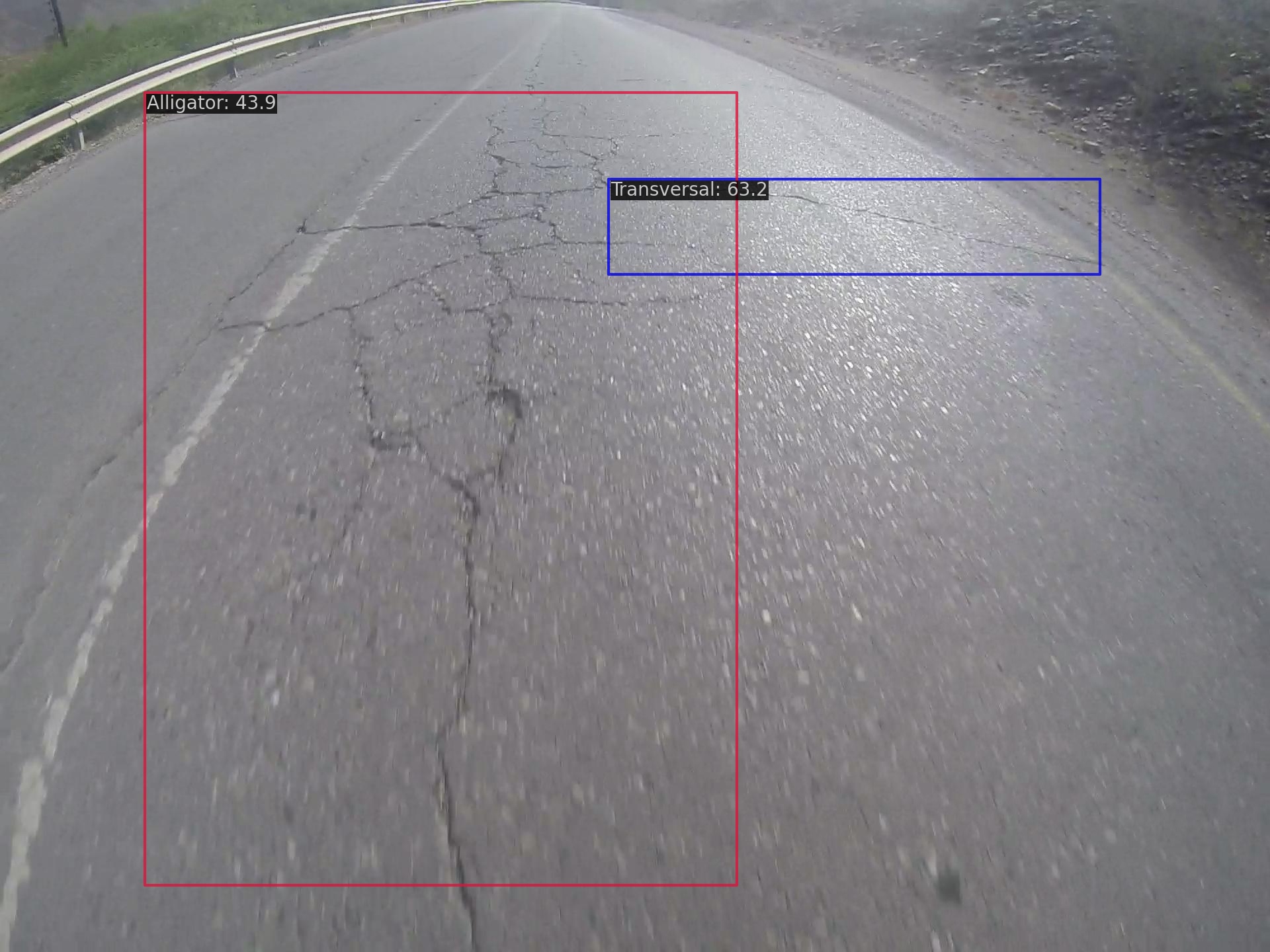}&
  \includegraphics[width=.19\textwidth]{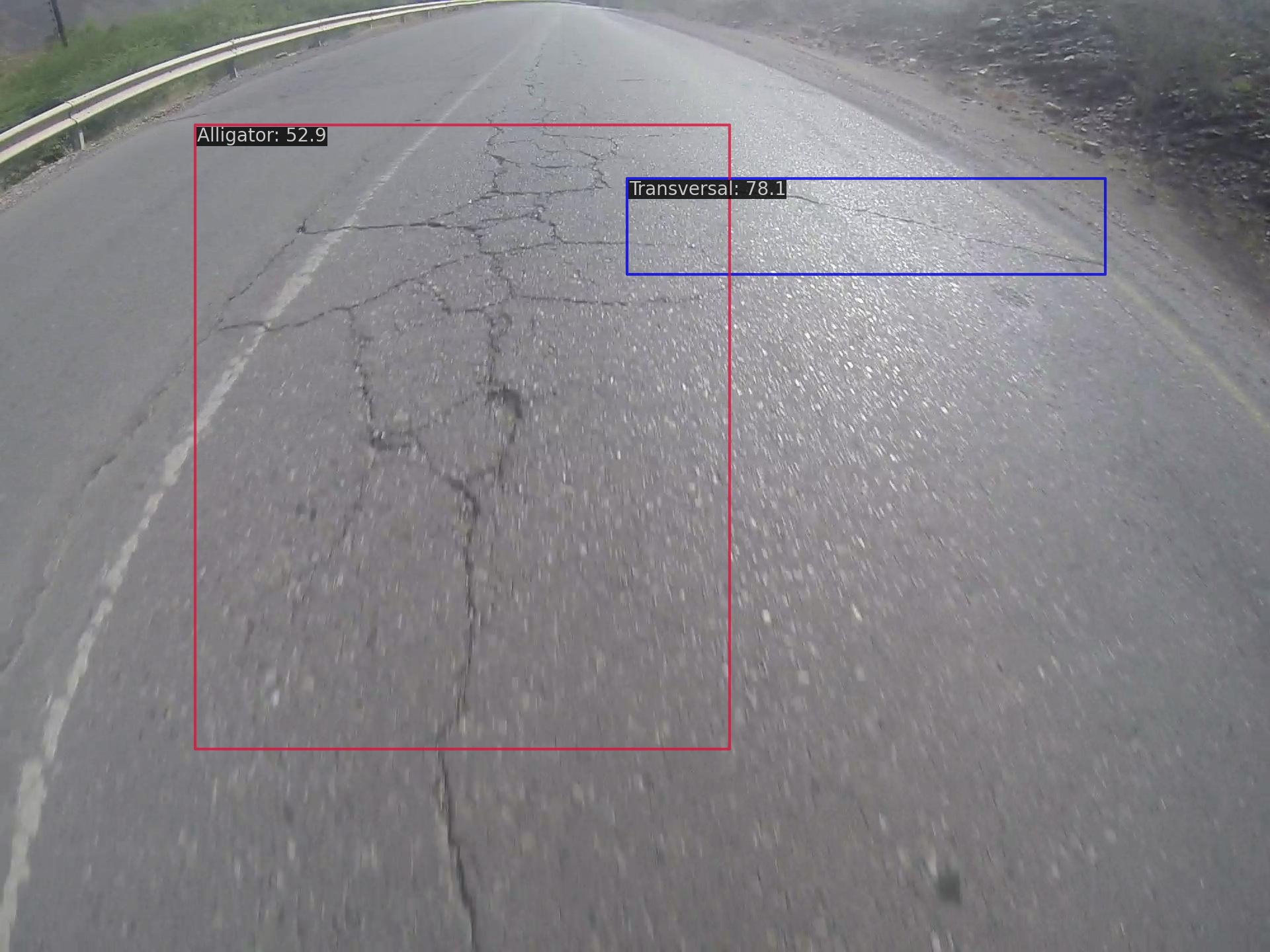}\\

  \includegraphics[width=.19\textwidth]{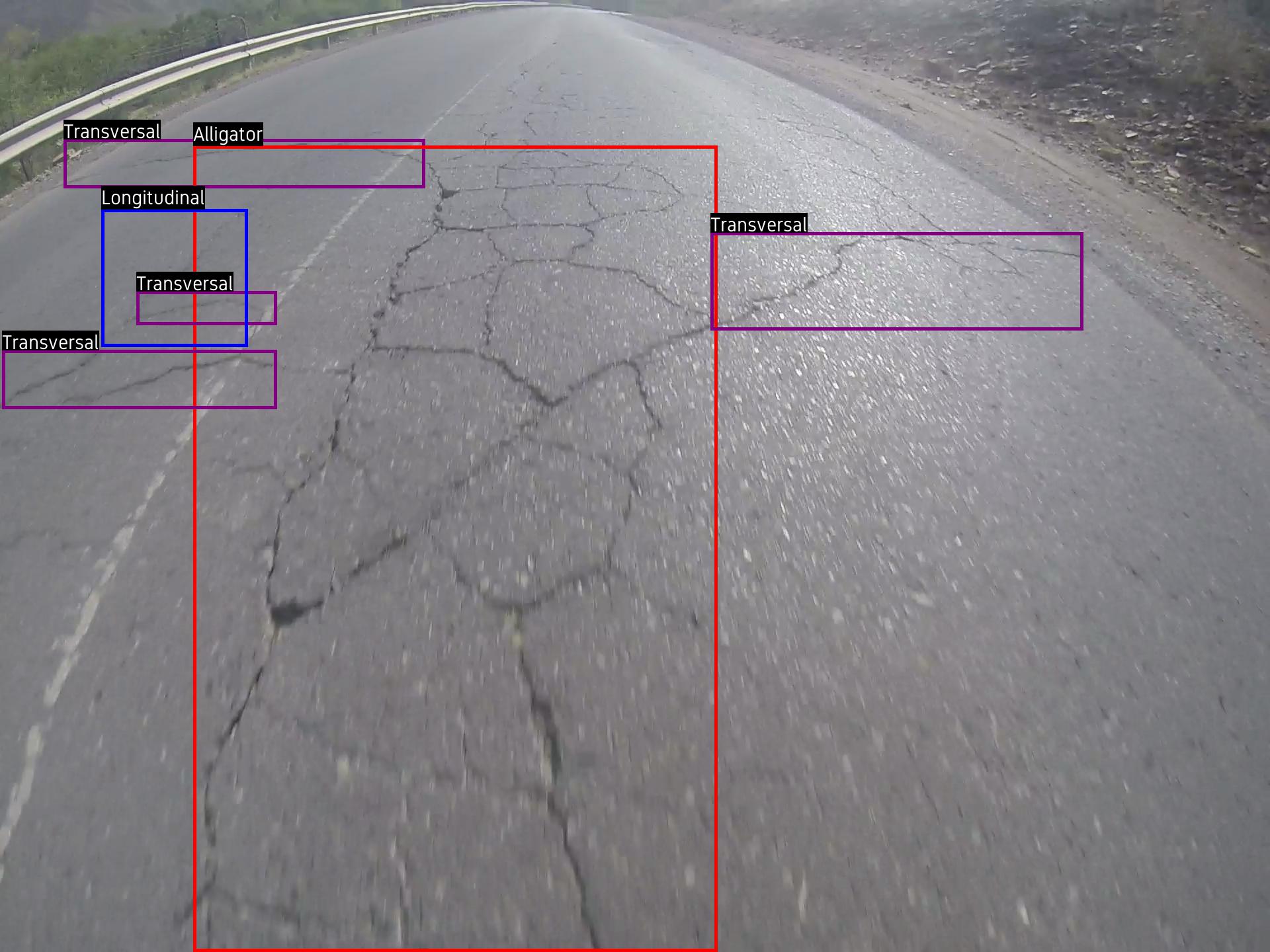}&
  \includegraphics[width=.19\textwidth]{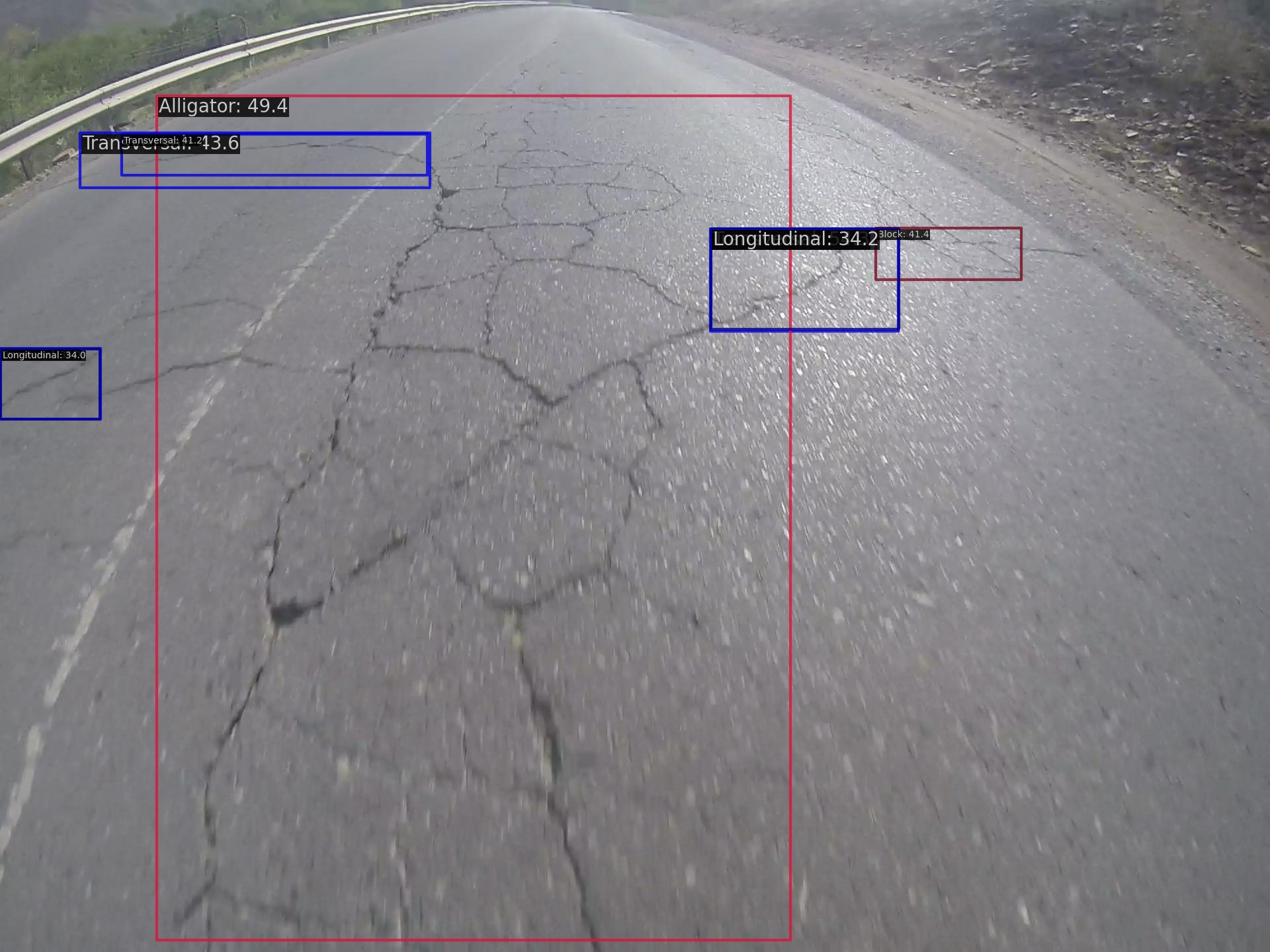}&
  \includegraphics[width=.19\textwidth]{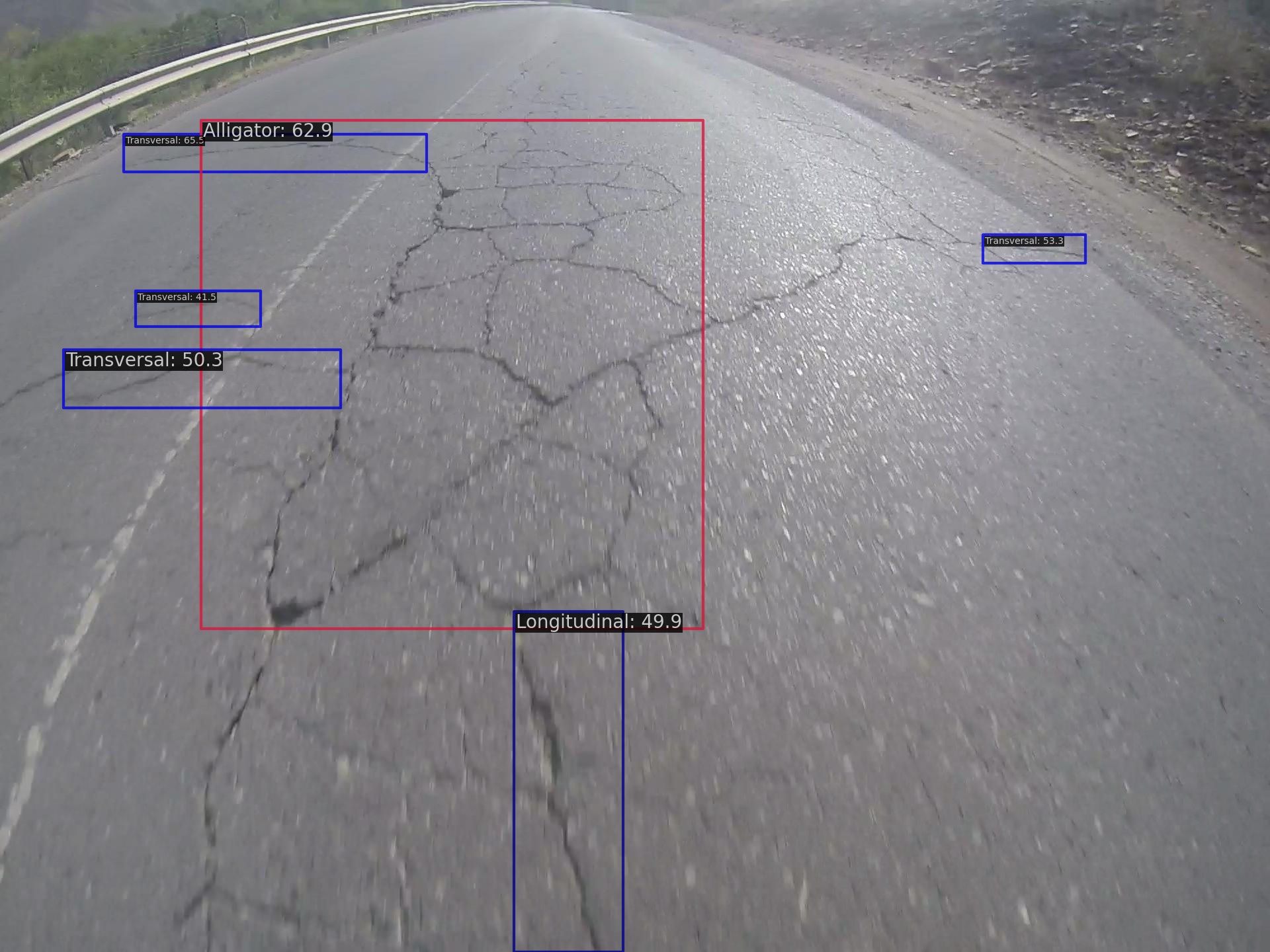}&
  \includegraphics[width=.19\textwidth]{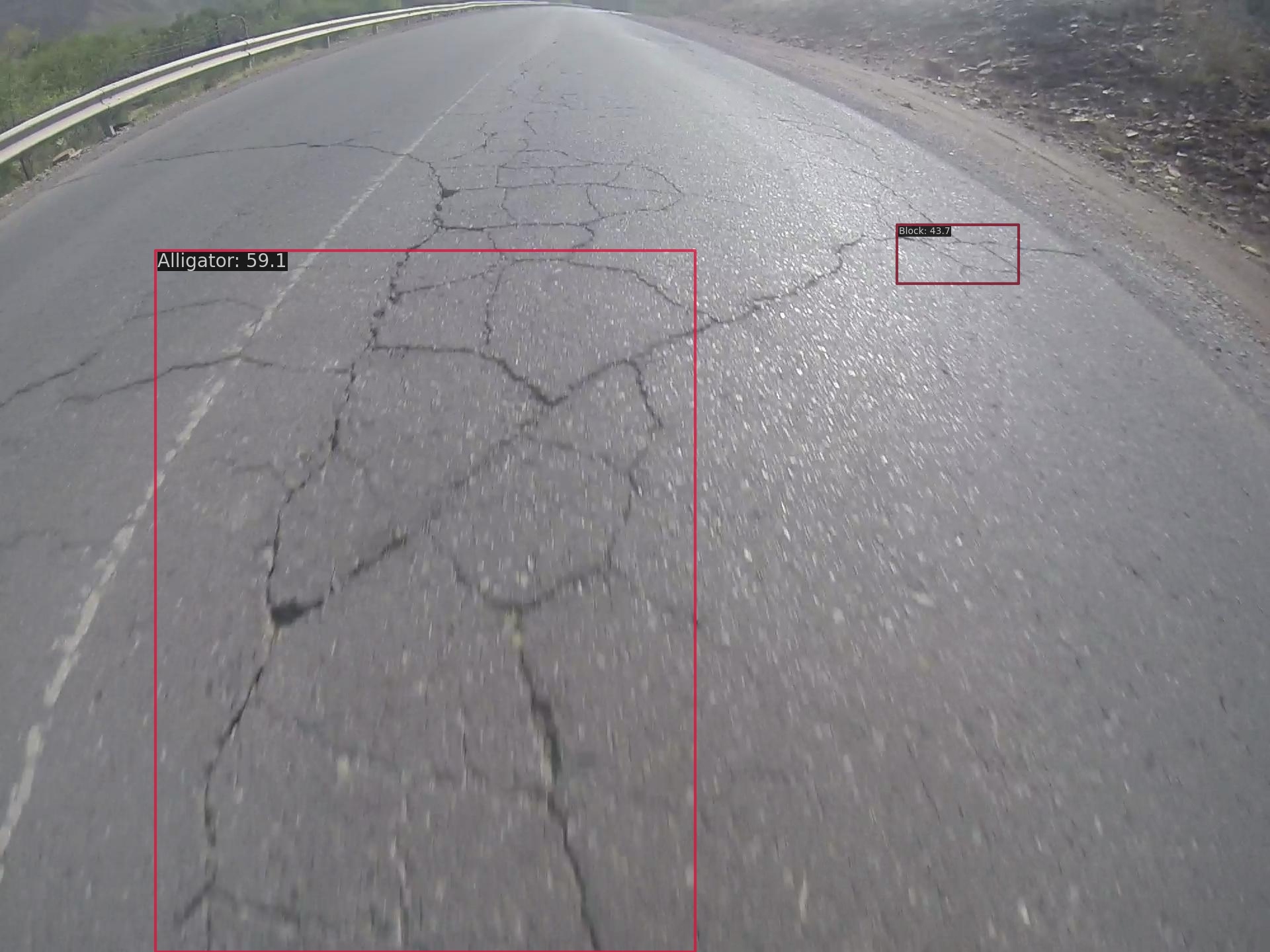}&  
  \includegraphics[width=.19\textwidth]{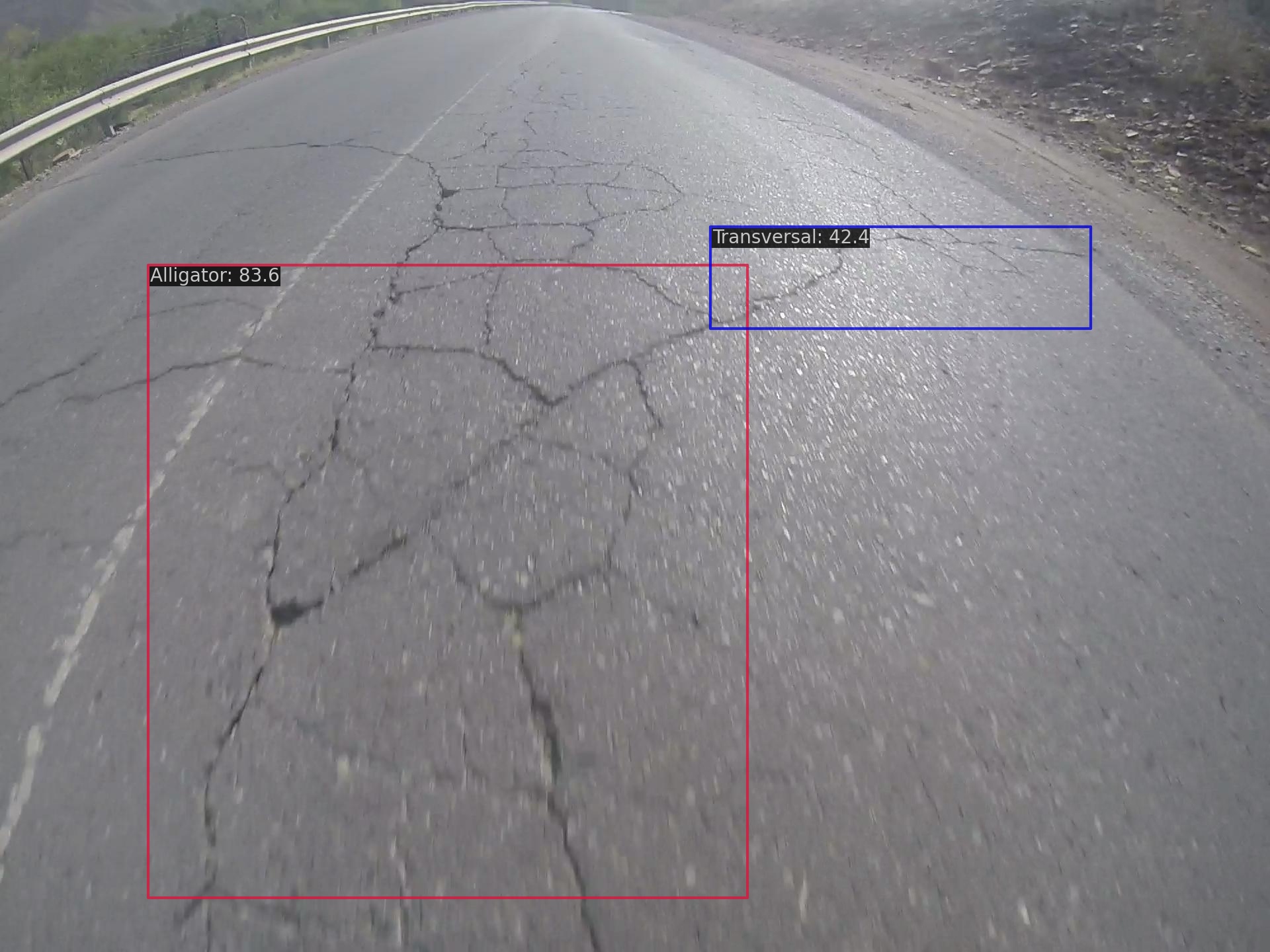}\\

   GT Annotations& RDD4D (Ours) & PPYOLOE & RTMDet & YOLOV8\\
\end{tabular}
\end{center}
\caption{Qualitative comparison of road damage detection using different object detection techniques. The columns present (from left to right) ground-truth annotations, our proposed approach, RTMDet, and YOLOv8, applied to four representative test images. Each row shows the same input image processed by these techniques, enabling direct comparison of detection accuracy and precision. The results demonstrate that our proposed methodology achieves more accurate detection and classification of road damage categories compared to the baseline RTMDet and other state-of-the-art approaches.}
\label{fig:mixup_images}
\end{figure*}

\section{Methodology}~\label{Methodology}
\begin{figure*}[t]
	\centering
	\includegraphics[width=0.90\textwidth]{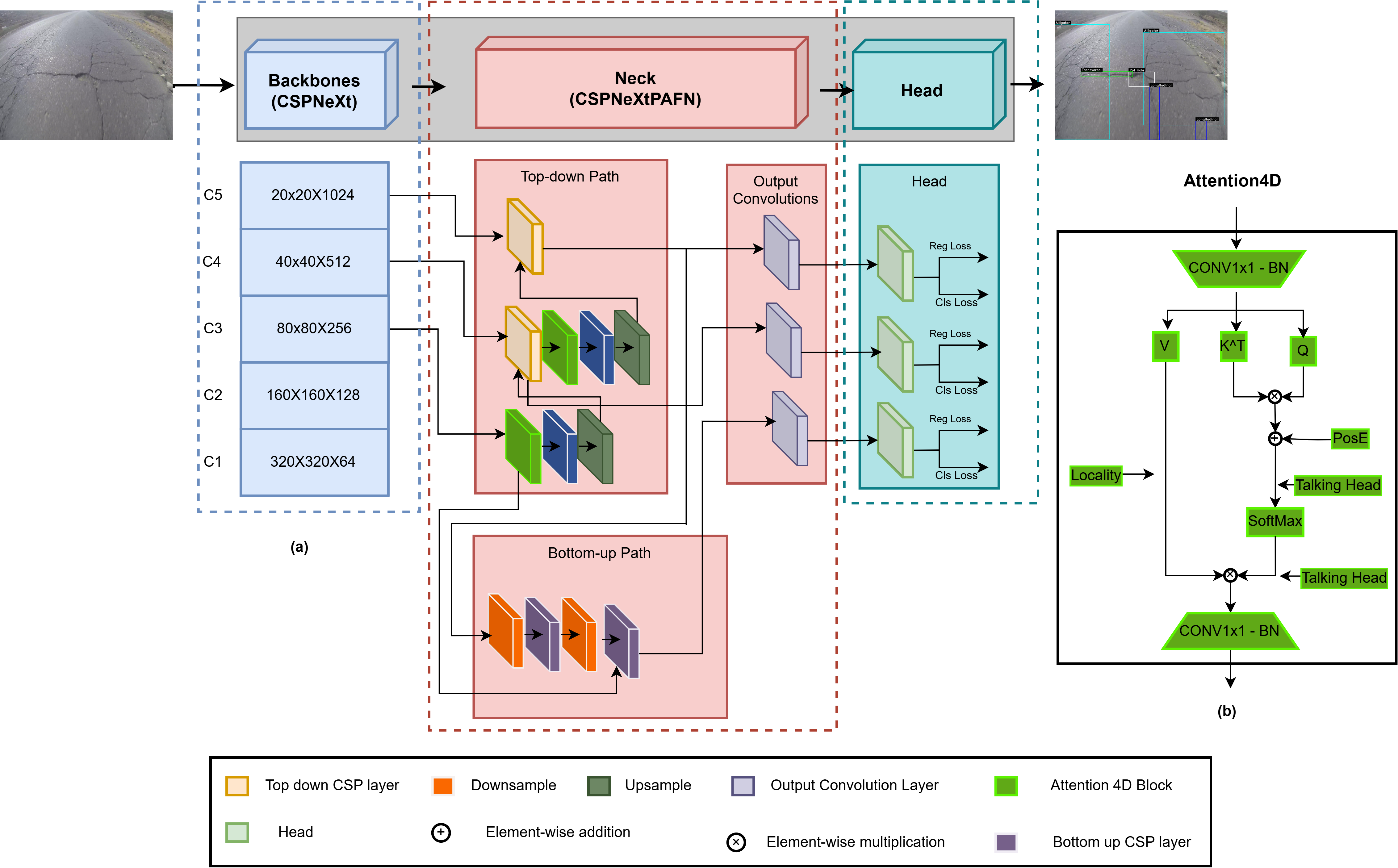}
	\caption{\textbf{Our proposed architecture:} The flow of feature maps through the neck section of the object detection network. Attention4D blocks (highlighted in bold green) are strategically applied to the feature maps from the backbone, enhancing spatial and channel-wise information. The neck processes these enhanced features through top-down and bottom-up paths, creating a multi-scale feature representation. }
	\label{proposed_methodology}
\end{figure*}

Due to its superior performance on the COCO dataset and its compatibility with edge computing devices, we chose the Real-Time Model Detection (RTMDet) lightweight object detection model as our baseline. This model is the latest addition to the one-stage real-time object detection family, and it has achieved outstanding performance in object detection and instance segmentation tasks. As illustrated in Figure~\ref{proposed_methodology}(a), the RTMDet architecture comprises three main blocks: the backbone, neck, and head. Recent studies typically adopted CSPDarkNet~\cite{bochkovskiy2020yolov4} as a backbone architecture in the RTMDet; however, we have employed CSPNeXt, an architecture that combines the ResNeXt Network and the Cross Stage Partial Network (CSP), as our backbone network~\cite{guo2021beyond} as the architecture utilizes large-kernel, depth-wise separable convolutions in the base modules. Our choice is based on the observation that the model quickly learns the global context by making the network deeper and wider.

The backbone model is available in two options. The first choice is the smaller architecture, denoted as P$_5$, while the second version, P$_6$, is the larger variant. We have opted for the former version, which produces five feature map scales, denoted as C$_1$, C$_2$, C$_3$, C$_4$, and C$_5$, after downsampling by factors of 2$^k$, where $k \in \{1, 2, \cdots, 5 \}$. The Neck module performs feature fusion using the final three feature map scales of C$_3$ (i.e., 8, 256, 80, 80), C$_4$ (i.e., 8, 512, 40, 40), and C$_5$ (i.e., 8, 1024, 20, 20). Moreover, the neck module further incorporates a multiscale feature pyramid network (CSPNeXtPAFPN) using the three feature maps from the backbone with both top-down and bottom-up feature propagation~\cite{lin2017feature,liu2018path} for enhancing the features before passing it into the head for detection and classification.

It should be noted that the CSPNeXtPAFPN architecture's top-down and bottom-up pathways, though effective, may only partially exploit the probable relations between features at different scales, potentially leading to a loss of fine-grained details. The neck part of RTMDet is responsible for fusing and refining multiscale features from the backbone. Hence, we propose and integrate Attention4D blocks. By exploiting this, the mentioned model can more effectively capture and utilize local and global contextual information across different scales. Eventually, the feature map of each scale is then employed by the detection head to anticipate object-bounding boxes and their categories. This architecture works well for general and rotated objects. It can also be extended for segmentation by adding kernel and mask feature generation heads~\cite{tian2020conditional}.

\noindent\vspace{1mm}\textbf{Attention4D Module}: The core innovation in our model is the Attention4D block, depicted in Figure~\ref{proposed_methodology}(b). This block processes input feature maps through a series of operations. A 1$\times$1 convolution is followed by batch normalization to adjust the channel dimension. The output is split into three branches: Value (V), Transposed Key $K^T$, and Query (Q). The Key and Query branches undergo element-wise multiplication. The result combines positional encoding (PosE) and a ``Talking Head" input. A softmax operation normalizes the attention weights. The softmax output is multiplied element-wise by the value branch and another ``Talking Head" input. Finally, the result passes through another 1$\times$1 convolution and batch normalization. This Attention4D block allows the network to capture local and global contextual information, potentially improving the model's understanding of complex scenes. The Attention4D blocks are strategically integrated at crucial points in both top-down paths of the neck architecture.

In the top-down path, an Attention4D block processes the input from the highest-level feature map before further refinement.
We apply a second Attention4D block after the first top-down CSPLayer. The bottom-up path feeds the output of the Attention4D block from the top-down path into a downsample operation before it reaches the bottom-up CSPLayers. This integration allows the network to refine features using our attention mechanism at multiple scales, potentially capturing fine-grained details and high-level semantic information.

\noindent\vspace{1mm}\textbf{Loss Function:}
A crucial step in training one-stage object detectors is matching dense predictions across various scales with the ground truth bounding boxes. This process, known as label assignment, has evolved over time by developing several strategies~\cite{feng2021tood,lin2017focal,tian2022fully}. While recent approaches have focused on dynamic methodologies for label assignment~\cite{carion2020end,ge2021ota,ge2021yolox}, we have adopted the dynamic soft label assignment strategy~\cite{lyu2022rtmdet}, which uses cost functions aligned with the training loss as the matching criteria to assign labels to the predicted bounding boxes more accurately. This dynamic soft label assignment strategy is based on the SOTA framework~\cite{ge2021yolox}. The cost function is defined as follows:
\begin{equation}
    \ell = \lambda_1 \cdot \delta + \lambda_2  \cdot \theta  + \lambda_3  \cdot \rho,
    \label{eq:totalloss}
\end{equation}
where \(\lambda_1 = 1\), \(\lambda_2 = 3\), and \(\lambda_3 = 1\) are the default weighting factors. Furthermore, $\delta$, $\theta$, $\rho$, and $\ell$ represent the classification, location, center proximity, and overall costs, respectively. In traditional approaches, the $\delta$ often relies on binary labels. This can lead to scenarios where predictions with high classification scores but inaccurate bounding boxes receive low classification costs and vice versa. To address this, a soft label $y$ assignment strategy is proposed instead of binary labels. We calculate the modified classification cost as follows:

\begin{table}[!t]
\centering
\caption{Performance comparison of different methods (AP)}
\begin{tabular}{l|c|c|c|c}
\hline
\rowcolor{gray!50}\textbf{Methods} & \textbf{Alligator} & \textbf{Block} & \textbf{Longitudinal} & \textbf{Transversal}   \\
\hline\hline
\rowcolor{gray!10}YOLOV8~\cite{yolov8_ultralytics} &0.122  &0.050 & 0.242 & 0.145  \\
YOLOV7~\cite{wang2022yolov7} & 0.172 &0.200  &0.380
  &0.267    \\
\rowcolor{gray!10}YOLOV6~\cite{wang2022yolov7} &0.041  &0.003 &
 0.190 &0.205  \\
PPYOLOE~\cite{Xu2022PPYOLOEAE} &0.012  &0.100  &0.206  &0.131   \\
\rowcolor{gray!10}RTMDET~\cite{lyu2022rtmdet} &0.127  &0.111 & \textcolor{blue}{\underline{0.379}} & \textcolor{blue}{\underline{0.350}}    \\
YOLOX~\cite{yolox2021}  &\textcolor{blue}{\underline{0.143}}& \textcolor{blue}{\underline{0.400}} &0.255  &0.000     \\
\hline

\rowcolor{gray!10}\textbf{RDD4D (Ours)} &\textcolor{red}{\textbf{0.145}}
 &\textcolor{red}{\textbf{0.900}} &\textcolor{red}{\textbf{0.387}} &\textcolor{red}{\textbf{0.352}}  \\
\hline

\end{tabular}
\label{tab:ap_results}
\end{table}

\begin{equation}
\delta = \text{CE}(\hat{y}, y) \cdot (y - \hat{y})^2,
\end{equation}
where CE stands for cross entropy and $\hat{y}$ is the prediction. The dynamic aspect of this approach means that label assignments can change during training, allowing the model to focus on the most informative examples. SoftLabel, inspired by GFL~\cite{li2022generalized}, uses intersection-over-union (IoU) as a soft label to provide a more detailed evaluation of prediction accuracy. For the $\theta$, existing methods using generalized IoU~\cite{rezatofighi2019generalized} often lack sufficient differentiation between high- and low-quality matches. The soft label approach uses a logarithmic IoU scale $\theta = -\log(\text{IoU})$, amplifying the differences between good and poor matches, especially for lower IoU values. This helps the model distinguish between high- and low-quality predictions during training.


Finally, the center proximity cost replaces rigid center-based criteria~\cite{ge2021ota,ge2021yolox,zhang2020bridging} with the following formula:

\begin{equation}
\rho = \frac{\eta}{|\hat{y_c}- y_c| - \epsilon},
\end{equation}
where $\eta$ and $\epsilon$ are adjustable parameters, while $\hat{y_c}$ is the predicted center and $y_c$ is the actual center. In the overall cost function shown in Eq.~\ref{eq:totalloss}, the center cost encourages predictions that are not only accurate in terms of classification and bounding box regression but are also well-centered on the object. For example, if two predictions have similar classification confidence and bounding box overlap, the prediction closer to the center of the ground truth object will be favored during training.

\section{Experiments}~\label{experiments}
\subsection{Settings}

\begin{figure*}[tbp]

\centering
\begin{tabular}{c@{ }c@{ }c}

     \includegraphics[width=0.33\textwidth]{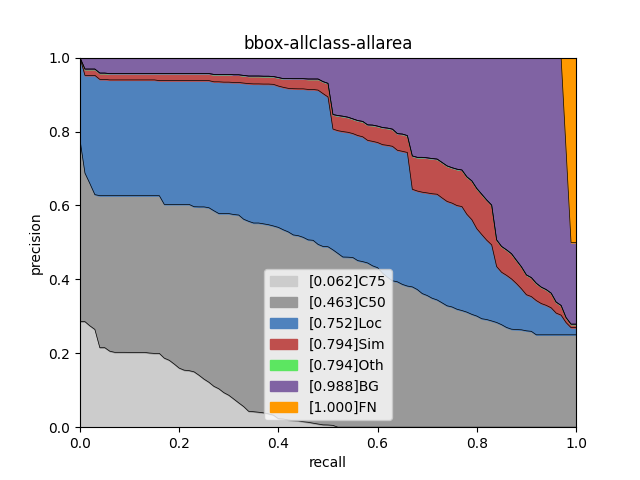}& 
     \includegraphics[width=0.33\textwidth]{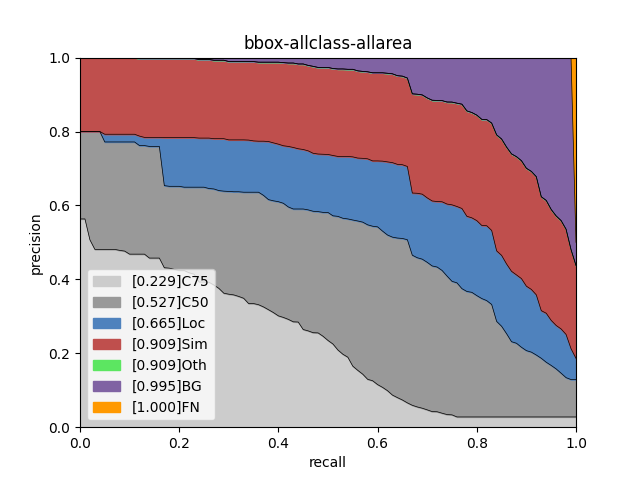}&
    \includegraphics[width=0.33\textwidth]{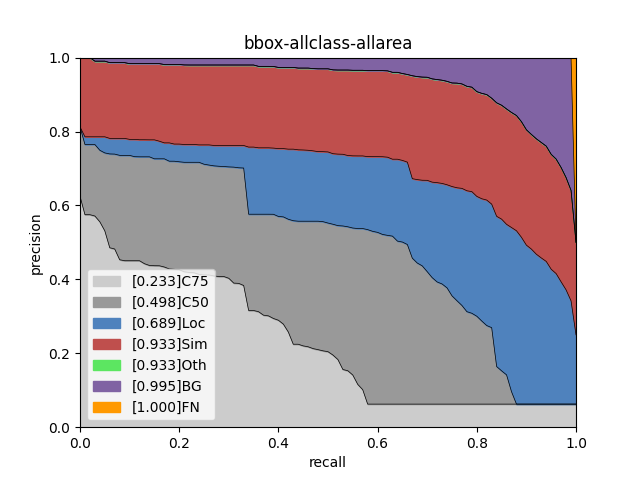}\\
     a) PPYOLOE &     b) RTMDET &     c) YOLOV7 \\
     \includegraphics[width=0.33\textwidth]{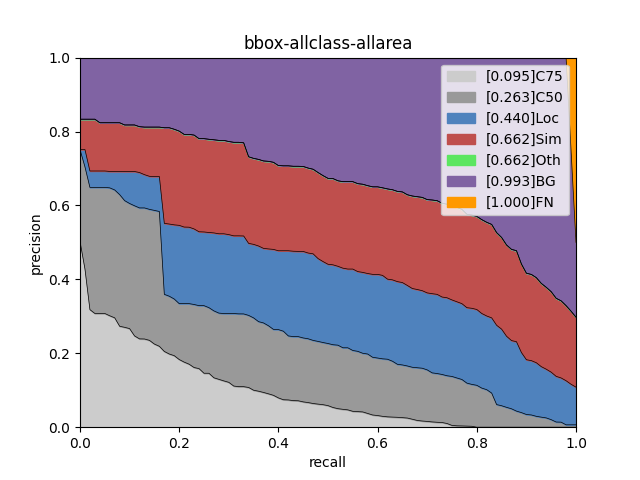}&
     \includegraphics[width=0.33\textwidth]{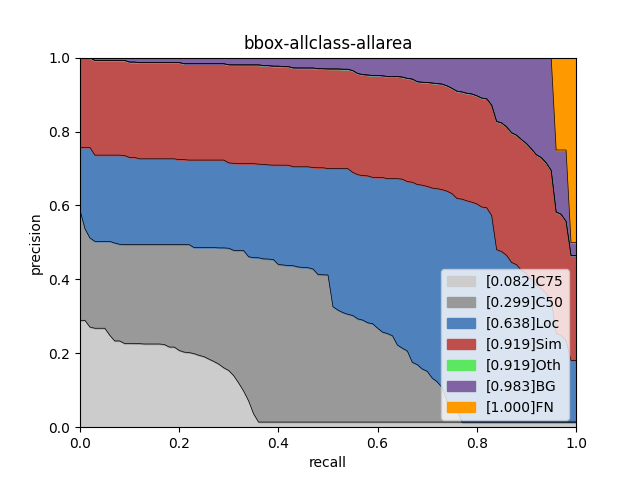}& 
     \includegraphics[width=0.33\textwidth]{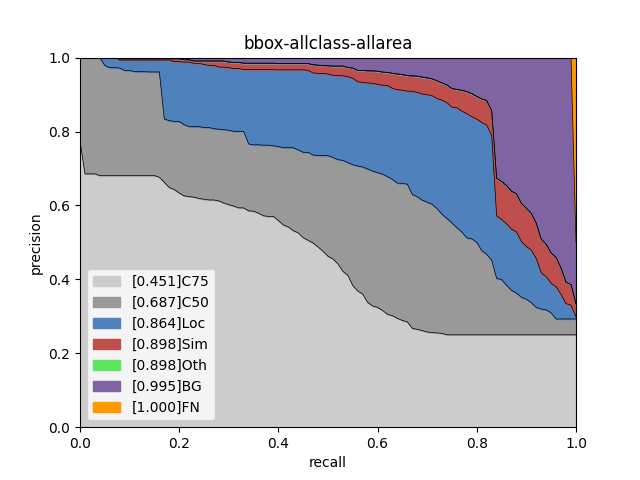}\\
     d) YOLOV6 &     e) YOLOV8 &     f) RDD4D (Ours)\\ 
    
     
\end{tabular}

\caption{Precision-recall curves comparing different methods for bounding box prediction across all classes in terms of all-area performance on a) PPYOLOE b) RTMDET c) YOLOV7 d) YOLOV6 e) YOLOV8 and f) Ours. The error analysis methods include C75 (gray), C50 (light gray), Loc (blue), Sim (red), Oth (green), BG (purple), and FN (orange), with their respective average precision scores shown in brackets. (Best viewed in color)}
\label{Figure7}
\end{figure*}





     


\begin{table*}[!t]
\centering
\caption{Scale-specific performance comparison (AP$_\text{S}$/AP$_\text{M}$/AP$_\text{L}$)}
\resizebox{\textwidth}{!}{
\begin{tabular}{l|ccc|ccc|ccc|ccc}
\hline
\multirow{2}{*}{\textbf{Methods}} & \multicolumn{3}{c|}{\textbf{Alligator}} & \multicolumn{3}{c|}{\textbf{Block}} & \multicolumn{3}{c|}{\textbf{Longitudinal}} & \multicolumn{3}{c}{\textbf{Transversal}} \\
\cline{2-13}

& S & M & L & S & M & L & S & M & L & S & M & L \\
\hline\hline
YOLOV8~\cite{yolov8_ultralytics} &-1.000  & -1.000 & 0.052 &-1.000 &-1.000  &0.053  & -1.000 & 0.076 & 0.252 & -1.000 & 0.091 &0.153 \\
YOLOV7~\cite{wang2022yolov7}  &-1.000  &-1.000  & 0.172 & -1.000 & -1.000 &0.200  & -1.000 &0.105  &0.397  &-1.000  & \textcolor{blue}{\underline{0.149}} &0.281   \\
YOLOV6~\cite{lyu2022rtmdet}  &-1.000 &-1.000  & 0.041&-1.000  &-1.000 &0.003  &-1.000 &0.062  &0.200  &-1.000 & \textcolor{red}{\textbf{0.153}} &0.214    \\
PPYOLOE~\cite{Xu2022PPYOLOEAE} & -1.000 &-1.000  & 0.012 &-1.000  &-1.000  & 0.023 &-1.000  &-1.000  &0.100 &-1.000  & 0.078 & 0.215   \\
RTMDET~\cite{lyu2022rtmdet}  &-1.000 & -1.000 &0.127 & -1.000 & -1.000&0.113  &-1.000 &\textcolor{blue}{\underline{0.124}}  &\textcolor{blue}{\underline{0.420}}  &-1.000 &  0.122&\textcolor{blue}{\underline{0.375}}   \\
YOLOX~\cite{yolox2021} & -1.000 & -1.000 &\textcolor{blue}{\underline{0.143}}  &-1.000  &-1.000  &\textcolor{blue}{\underline{0.400}}  & -1.000 & 0.011 &0.271  &-1.000  &0.000  &0.000    \\
\hline 
\rowcolor{gray!10}\textbf{RDD4D (Ours)} &-1.000 & -1.000&\textcolor{red}{\textbf{0.145}}&-1.000&-1.000 &\textcolor{red}{\textbf{0.900}} &-1.000 & \textcolor{red}{\textbf{0.272}}&\textcolor{red}{\textbf{0.699}} &-1.000& 0.132& \textcolor{red}{\textbf{0.379}}\\
\hline  
\end{tabular}}
\label{tab:scale_results}
\end{table*}

\noindent 
\textit{Training Details.}
During training, we used a batch size of four in our experiments. We set the weight decay at 0.9, suggesting a considerable regularization impact to counteract overfitting, although this number is more significant than usual. Compared to regular values, which typically range between 0.8 and 0.99, we set the momentum at 5e$^{-4}$. We train the model for 300 epochs to ensure complete performance optimization.

\begin{table*}[tbp]
\footnotesize
\caption{Detection results (mAP) on road-crack dataset. The best results are shown in \textcolor{red}{\textbf{bold}} and the second best \textcolor{blue}{\underline{underlined}}.}

\centering
\begin{tabular}{l|l|c|c|c|c|c|c|c|c|c|c} \hline
\rowcolor{gray!50}Methods & Backbone  & AP & AP\textsubscript{50} & AP\textsubscript{75} &AP\textsubscript{S}  &AP\textsubscript{M} &AP\textsubscript{L} &AR &AR\textsubscript{S}  &AR\textsubscript{M} &AR\textsubscript{L} \\ \hline \hline 
YOLOV8~\cite{yolov8_ultralytics} & YOLOv8CSPDarknet & 0.122 & 0.299 & 0.082 & -1.000 & 0.083 & 0.127 & 0.448 & -1.000 & 0.234 & 0.454\\
       
\rowcolor{gray!10}YOLOV7~\cite{wang2022yolov7}  & YOLOv7Backbone & \textcolor{blue}{\underline{0.255}} & \textcolor{blue}{\underline{0.498}} & \textcolor{blue}{\underline{0.233}} & -1.000 & \textcolor{red}{\textbf{0.127}} & 0.263 & 0.547 & -1.000 & 0.351 & 0.553 \\

\rowcolor{gray!10}YOLOV6~\cite{wang2022yolov7}  & YOLOv6Backbone &0.110  &0.263  & 0.095 &-1.000  &0.108  & 0.114 & \textcolor{blue}{\underline{0.560}} & -1.000 & \textcolor{red}{\textbf{0.460}} &0.572  \\

PPYOLOE~\cite{Xu2022PPYOLOEAE}  & PPYOLOECSPResNet & 0.112 & 0.463 & 0.062 & -1.000 & 0.079 & 0.117 & 0.322 & -1.000 & \textcolor{blue}{\underline{0.388}} & 0.325\\
       
\rowcolor{gray!10}RTMDET~\cite{lyu2022rtmdet}  & CSPNeXt  & 0.268 & 0.527 & 0.229 & -1.000 & \textcolor{blue}{\underline{0.123}} & \textcolor{blue}{\underline{0.280}} & 0.517 & -1.000 & 0.373 & \textcolor{blue}{\underline{0.623}}\\ 

YOLOX~\cite{yolox2021} & YOLOXCSPDarknet &  0.200&0.377 &0.188 &-1.000 &0.006 &0.204 &0.288 &-1.000 &0.033 &0.386\\

\hline 
\rowcolor{gray!10}\textbf{Ours} &CSPNeXt &\textcolor{red}{\textbf{0.446}} &\textcolor{red}{\textbf{0.687}} &\textcolor{red}{\textbf{0.451}}&-1.000&0.113&\textcolor{red}{\textbf{0.458}}&\textcolor{red}{\textbf{0.675}}&-1.000&0.277&\textcolor{red}{\textbf{0.690}}\\
\hline  
\end{tabular}
\label{tab:det_results}
\end{table*}

\begin{figure}[tbp]
\begin{tikzpicture}
\begin{axis}[
bar width=10,
ybar,
enlargelimits=0.12,
legend style={at={(0.2,0.945)},
anchor=north,legend columns=1},
ylabel={Class Area (Pixcel Square)},
symbolic x coords={Alligator,Block,Longitudinal, Traversal, Pothole},
xtick=data,
nodes near coords,
nodes near coords align={vertical},
nodes near coords style={font=\tiny, inner xsep=0.5pt},
x tick label style={rotate=45,anchor=east},
]
\addplot coordinates {(Alligator,0) (Block,0) (Longitudinal,0) (Traversal,1) (Pothole,0)};
\addplot coordinates {(Alligator,1) (Block,4) (Longitudinal,74) (Traversal,232) (Pothole,6)};
\addplot coordinates {(Alligator,65) (Block,1) (Longitudinal,1579) (Traversal,1936) (Pothole,15)};
\legend{Small, medium, Large}
\end{axis}
\end{tikzpicture}
	\caption{The graph presents a ground-truth analysis of bounding box areas, where each bar represents the count of annotations. These annotations are categorized into three distinct groups—small, medium, and large—based on the size of their bounding box areas.}
	\label{ground_truth_analysis}
\end{figure}
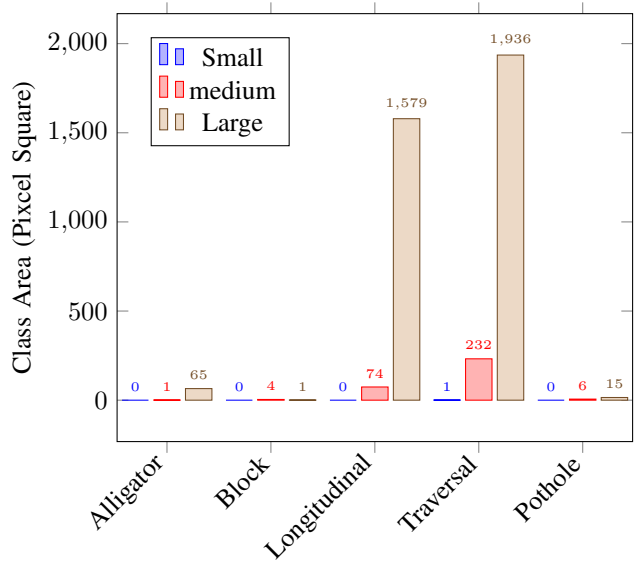

\vspace{2mm}
\noindent 
\textit{Datasets.} 
The road crack dataset videos were recorded using a GoPro camera mounted on the front of a car and aimed at the road surface. As the vehicle traversed various roads, it recorded multiple types of cracks under varying lighting conditions, comprehensively representing different road conditions. A civil engineer carefully reviewed and validated each road condition to accurately identify and confirm the presence of various types of cracks. We recorded 30 video segments in MP4 format, each with a resolution of 1920$\times$1440 pixels and a frame rate of 30fps. Most video segments lasted between 20 to 30 minutes. These videos were then converted into frames, with blurry images due to car speeds discarded to maintain quality. We have used the Computer Vision Annotations Tool to annotate images in Pascal-VOC and COCO formats. From the video footage, we annotated a total of 1,500 images. Figure~\ref{ground_truth_analysis} presents the distribution of the dataset. Some video frames contained up to 45 annotations, showing the complex and cluttered nature of various road cracks within a single frame. The COCO metrics categorize the Objects in the dataset into three sizes based on the area of their bounding boxes. Small objects occupy less than 32$\times$32 pixels, medium objects are between 32$\times$32 pixels and 96$\times$96 pixels, and large objects exceed 96$\times$96 pixels. According to our ground-truth analysis, as detailed in Figure~\ref{ground_truth_analysis}, most annotated objects are classified as large or medium based on COCO metrics.

\vspace{2mm}
\noindent 
\textit{Evaluation Metrics.}
To evaluate the performance of our object detectors, we have used the COCO evaluation metrics, a standard metric for assessing both object detection and segmentation models. The COCO evaluation framework provides a detailed analysis of model performance across various Intersection over Union (IoU) thresholds and categorizes object detection results based on different object sizes—small, medium, and large. 

\section{Comparisons}

\subsection{Comparisons on our proposed DRDD dataset}
Precision and recall are essential metrics used to evaluate the performance of a model. Precision measures the accuracy of the detections, which is the proportion of predicted positive identifications that were actually correct. A high precision rate indicates that the model made few mistakes in its identifications, making it reliable when it predicts that an object is present. On the other hand, recall assesses the model's ability to capture all actual positive detections within the dataset.  It reflects the proportion of true positive detections made by the model relative to the total number of positives that actually exist in the data. 
High recall indicates that the model leaves few actual positives undetected, maximizing the chances that if an object is present, the model will recognize it. A balanced performance in both metrics is crucial for reliable road damage detection systems. We evaluate the performance of various deep learning models, i.e., YOLOv8~\cite{jocher2023ultralytics}, YOLOv7~\cite{wang2022yolov7}, YOLOV6~\cite{li2022yolov6} PPYOLOE~\cite{Xu2022PPYOLOEAE}, YOLOX~\cite{ge2021yolox}, and RTMDet~\cite{lyu2022rtmdet}, in detecting small, medium, and large road cracks using the metrics of Average Precision (AP) and Average Recall (AR) as shown in Table~\ref{tab:det_results}.

The AP and AR scores of -1.000 for small-sized damages (AP\textsubscript{S} and AR\textsubscript{S}) across all models indicates the absence of small damages in the dataset, rather than poor detection performance. This observation confirms that the majority of road damages in our dataset belong to the large category. For medium-sized damages, YOLOv7 achieved the highest AP of 0.127, followed by RTMDET with 0.123. In the large damage category, where most of our dataset's instances are concentrated, our RDD4D method significantly outperformed others with an AP of 0.458, while RTMDET showed the second-best performance with 0.280.

In terms of recall performance, YOLOV6 demonstrated strong capability in detecting medium-sized damages with the highest AR of 0.460, followed by PPYOLOE with 0.388. For large damages, which constitute the majority of our dataset, our method achieved the best AR of 0.690, with RTMDET following at 0.623. Overall, our proposed method demonstrates superior performance with the highest AP (0.446), AP50 (0.687), and AP75 (0.451) scores across all models, showing significant improvements over the second-best performer RTMDET (AP=0.268, AP50=0.527, AP75=0.229). These results validate the effectiveness of our approach, particularly in detecting the predominant large-sized road damages.

Regarding recall, all models struggled with small objects, scoring an AR of -1.000. This is due to the absence of small cracks in the dataset. PPYOLOE recorded the highest AR of 0.388 for medium-sized objects, indicating its effectiveness in recognizing the presence of medium-sized cracks more consistently than its counterparts despite its lower precision. Our model demonstrated the highest recall for large road cracks with an AR of 0.690, confirming its robustness in terms of precision and its ability to reliably identify a higher proportion of large road cracks present in the dataset.

A comprehensive analysis of various error distributions between our model and other models is shown in Fig.~\ref{Figure7}. Detection errors in road damage assessment can be categorized into several types: localization errors (Loc) when bounding boxes are imprecisely placed, similarity-based (Sim) confusions between visually similar damage types, category confusion (Oth) where damages are misclassified, background false positives (BG) where normal road surfaces are mistakenly identified as damage, and false negatives (FN) where actual damages go undetected. Our model achieves higher confidence scores (C75=0.451, C50=0.687) compared to YOLOV6 (C75=0.095, C50=0.263), YOLOV7 (C75=0.233, C50=0.498), and YOLOV8 (C75=0.082, C50=0.299). The localization accuracy (Loc=0.864) is superior to all other models, indicating more precise bounding box predictions. Our model maintains better precision across higher recall values, with a more balanced distribution of similarity-based and category confusion errors (both Sim and Oth=0.898). While background detection errors (BG=0.995) and false negatives (FN=1.000) are present, they occupy a relatively smaller portion of the graph, particularly at high recall values. The smoother curves and more gradual degradation in precision as recall increases suggest a more robust and stable detection performance. This comprehensive improvement across all error categories demonstrates that our model not only achieves better absolute performance but also handles the various types of detection challenges more effectively than existing approaches.

A classwise AP is given in Table~\ref{tab:ap_results} and detailed scale-specific analysis across different damage types is presented in Table \ref{tab:scale_results}. The value of -1.000 for small (S) and medium (M) sizes in Alligator and Block cracks, and small sizes in Transversal cracks indicates the absence of these scale variations in our dataset. Our method demonstrates superior performance in detecting large-scale damages across all categories, achieving the highest AP scores for Alligator (0.145), Block (0.900), Longitudinal (0.699), and Transversal cracks (0.379).

This comprehensive analysis confirms our model's robust performance across different damage types and scales, particularly excelling in the detection of large-scale road damages which constitute the majority of real-world cases.


\subsection{Comparisons on CrackTinyNet dataset}
Our proposed method demonstrates significant improvements over existing approaches on CrackTinyNet~\cite{CrackTinyNet} as shown in Table~\ref{tab:ComparisonRDD}. 
The dataset categorizes road surface defects into seven distinct classes: three crack types (Longitudinal-D00, Transverse-D10, and Alligator-D20), surface deformation (Pothole-D40), two types of marking deterioration (White Line-D43 and Cross Walk-D44 Blur), and infrastructure elements (Manhole Cover-D50).
With a MAP@.5 score of 0.825, our model substantially outperforms the previous best performer, CrTNet~\cite{CrackTinyNet}(0.601), on this dataset, representing a 37\% improvement in detection accuracy. This enhancement is particularly noteworthy when compared to widely used object detection frameworks like YOLOv8 ~\cite{jocher2023ultralytics}(0.445) and EfficientDet~\cite{tan2020efficientdet} (0.552).
Our model makes fewer false positive detections, crucial for real-world applications where false alarms could lead to unnecessary maintenance inspections. Traditional models like YOLOv5s~\cite{glenn_jocher_2022_7002879} and SSD~\cite{liu2016ssd} show considerably lower precision (0.27 and 0.31 respectively). Perhaps most impressively, our model achieves a recall rate of 0.98, demonstrating its ability to detect almost all instances of road damage in the dataset. This is a substantial improvement over CrTNet's 0.61 recall and far exceeds other benchmarks like Fast R-CNN~\cite{ren2016faster} (0.55) and YOLOv8 (0.53).

\begin{table}[tbp]
\centering
\caption{Road damage categories and their descriptions with class-wise Average Precision (AP) results for DRDD.}
\resizebox{\columnwidth}{!}{
\begin{tabular}{l|l|c|c}
\hline
\rowcolor{gray!50}\textbf{Class} & \textbf{Damage Type} & \textbf{AP@[IoU=0.50:0.95]} & \textbf{AP@[IoU=0.50]}  \\
\hline\hline
\rowcolor{gray!10}D00 & Longitudinal Crack & 0.829 & 0.833 \\
D10 & Transverse Crack & 0.916 & 0.928 \\
\rowcolor{gray!10}D20 &Aligator Crack & 0.863 & 0.863 \\
D40 &Pothole & 0.577 & 0.609 \\
\rowcolor{gray!10}D43 &White Line Blur & 0.689 & 0.697 \\
D44 & Cross Walk Blur & 0.944 & 0.945 \\
\rowcolor{gray!10}D50 &Manhole Cover & 0.853 & 0.903 \\
\hline
\end{tabular}}
\label{tab:ap_results_RDD}
\end{table}

\begin{figure}[t]
    \centering
    \includegraphics[width=1\columnwidth]{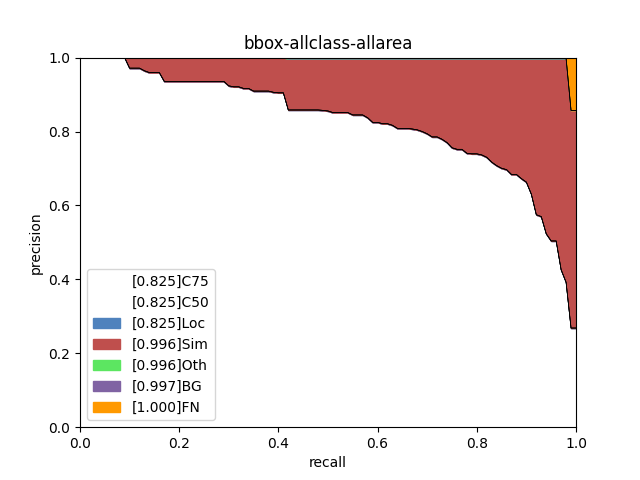}
    \caption{The graph demonstrates various error types affecting detection performance: localization accuracy at IoU thresholds of 0.75 (C75) and 0.50 (C50), localization errors (Loc), similarity-based false positives (Sim), other category confusions (Oth), background detections (BG), and false negatives (FN).}
    \label{fig:other_dataset_performance}
\end{figure}

The precision and recall graph in Figure~\ref{fig:other_dataset_performance} shows excellent model performance with uniform confidence scores (both C75 and C50 at 0.825) across different IoU thresholds. The dominant red region representing similarity errors (Sim=0.996) indicates that most detection errors stem from confusion between similar damage types rather than localization issues (Loc=0.825) or background misidentifications (BG=0.997). The model maintains high precision across most recall values, with performance degrading only at very high recall levels.
The class-wise performance analysis in Table \ref{tab:ap_results_RDD} shows that our model excels at detecting Cross Walk Blur (D44) and Transverse Crack (D10), achieving AP scores above 0.90 across both IoU thresholds. The detection of Aligator Crack (D20), Manhole Cover (D50), and Longitudinal Crack (D00) also demonstrates strong performance with AP scores above 0.82. However, the model shows relatively lower accuracy in detecting White Line Blur (D43) and Potholes (D40), with Potholes being the most challenging category at AP scores of 0.577 and 0.609, likely due to their irregular shapes and varying appearances.

\begin{table}[tbp]
\centering
\caption{Quantitative comparison of detection performance between our RDD4D and existing state-of-the-art methods evaluated on the CrackTinyNet~\cite{CrackTinyNet} dataset. The best results are shown in \textcolor{red}{\textbf{bold}} and the second best \textcolor{blue}{\underline{underlined}}.}
\begin{tabular}{l|c|c|c}
\hline
\rowcolor{gray!50}Name & MAP@.5 & Precision & Recall \\
\hline\hline
YOLOv5s & 0.381 & 0.27 & 0.77 \\
\rowcolor{gray!10}YOLOv5x & 0.404 & 0.45 & 0.51 \\
SSD & 0.406 & 0.31 & 0.58 \\
\rowcolor{gray!10}Fast R-CNN & 0.443 & 0.42 & 0.55 \\
YOLOv8 & 0.445 & 0.43 & 0.53 \\
\rowcolor{gray!10}EfficientDet & 0.552 & 0.38 & 0.69 \\
CrTNet & \textcolor{blue}{\underline{0.601}} & \textcolor{blue}{\underline{0.62}} & \textcolor{blue}{\underline{0.61}} \\
\hline
\rowcolor{gray!10}Ours &\textcolor{red}{\textbf{0.825}}   & \textcolor{red}{\textbf{0.81}}&\textcolor{red}{\textbf{0.98}} \\
\hline
\end{tabular}
\label{tab:ComparisonRDD}
\end{table}

\subsection{Ablation Studies}
The ablation study on Attention4D block placement strategies, as shown in Table \ref{tab:placement_strategies}, reveals that utilizing blocks in the top-down path yields superior performance (mAP=0.446, AP50=0.687, AP75=0.451) compared to bottom-up path (mAP=0.412) or single block placement (mAP=0.398). While implementing blocks in both paths marginally improves performance (mAP=0.455), it comes at the cost of reduced inference speed (25.1 FPS vs 26.8 FPS) and increased parameter count (40.2M vs 38.5M). This suggests that the top-down path placement provides the optimal balance between detection accuracy and computational efficiency.
Further analysis of the number of Attention4D blocks in Table \ref{tab:num_blocks} demonstrates a clear performance trend. While a single block shows modest results (mAP=0.425), implementing two blocks achieves significantly better performance (mAP=0.446) with a reasonable parameter increase of 1.3M. Although adding three or four blocks marginally improves accuracy (mAP=0.449 and 0.451 respectively), the computational overhead becomes substantial, reducing the FPS to 25.3 and 23.7 respectively, while increasing the parameter count to 39.8M and 41.1M. These results validate our choice of using two Attention4D blocks in the top-down path as the optimal configuration, providing the best trade-off between detection accuracy and computational efficiency.
\begin{table}[!t]
\centering
\caption{Performance comparison of different placement strategies in RDD4D. The best results are provided when both path are used.}
\begin{tabular}{l|c|c|c|c|c}
\hline
\rowcolor{gray!50}\textbf{Placement Strategy} & \textbf{mAP} & \textbf{AP50} & \textbf{AP75} & \textbf{FPS} & \textbf{Params(M)} \\
\hline\hline
\rowcolor{gray!10}Only Top-down Path & 0.446 & 0.687 & 0.451 & 26.8 & 38.5 \\
Only Bottom-up Path & 0.412 & 0.654 & 0.423 & 26.5 & 38.5 \\
\rowcolor{gray!10}Single Block at End & 0.398 & 0.642 & 0.412 & 27.2 & 37.8 \\
Both Paths & 0.455 & 0.693 & 0.460 & 25.1 & 40.2 \\
\hline
\end{tabular}
\label{tab:placement_strategies}
\end{table}

\begin{table}[!t]
\centering
\caption{Impact of number of Attention4D blocks on performance of RDD4D. Using more blocks increases performance.}
\resizebox{\columnwidth}{!}{
\begin{tabular}{c|c|c|c|c|c}
\hline
\rowcolor{gray!50}\textbf{Blocks (Attention4D)} & \textbf{mAP} & \textbf{AP50} & \textbf{AP75} & \textbf{FPS} & \textbf{Params(M)} \\
\hline\hline
\rowcolor{gray!10}1  & 0.425 & 0.668 & 0.432 & 27.5 & 37.2 \\
2   & 0.446 & 0.687 & 0.451 & 26.8 & 38.5 \\
\rowcolor{gray!10}3   & 0.449 & 0.689 & 0.453 & 25.3 & 39.8 \\
4   & 0.451 & 0.690 & 0.455 & 23.7 & 41.1 \\
\hline
\end{tabular}}
\label{tab:num_blocks}
\end{table}

\section{Conclusion}\label{conclusion}
In this paper, we addressed the challenging task of detecting multiple types of road damage types by introducing an enhanced architecture with an integrated Attention4D mechanism. We collected a novel dataset that captures diverse road damage types, with most of the images containing different damage types., serving as a valuable resource for future research in this field. We have proven that our approach outperforms both our proposed dataset and the CrackTinyNet dataset in terms of overall performance. Our model performs well because the 4D attention module effectively processes both local and global contextual information. This work establishes a strong foundation for future developments in automated road damage detection systems, which are crucial for efficient infrastructure maintenance.
\section*{Acknowledgment}
The research project was funded by the Ministry of Higher Education, Scientific Research and Innovation, Oman under contract number (BFP/RGP/ICT/21/406).

\bibliographystyle{IEEEtran}
\bibliography{refs}
\end{document}